\documentclass[10pt,twocolumn,letterpaper]{article}
\pdfoutput=1

\usepackage{cvpr}
\usepackage{times}
\usepackage{epsfig}
\usepackage{graphicx}
\usepackage{amsmath}
\usepackage{amssymb}
\usepackage{subcaption}

\usepackage{algpseudocode}
\usepackage{algorithm}
\usepackage{multirow}

\makeatletter
\renewcommand{\paragraph}{%
  \@startsection{paragraph}{4}%
  {\z@}{1ex \@plus 1ex \@minus .2ex}{-1em}%
  {\normalfont\normalsize\bfseries}%
}
\makeatother

\algtext*{EndIf}
\renewcommand{\algorithmicrequire}{\textbf{Input~~~}:}
\renewcommand{\algorithmicensure}{\textbf{Output}:}

\usepackage[pagebackref=true,breaklinks=true,letterpaper=true,colorlinks,bookmarks=false]{hyperref}

\cvprfinalcopy 

\def\cvprPaperID{****} 

\ifcvprfinal\fi
\begin{document}

\title{Learning Multi-Domain Convolutional Neural Networks for Visual Tracking}

\author{Hyeonseob Nam~~~~~~~~~~Bohyung Han\\
Dept. of Computer Science and Engineering, POSTECH, Korea\\
{\tt\small \{namhs09, bhhan\}@postech.ac.kr}
}

\maketitle

\begin{abstract}
We propose a novel visual tracking algorithm based on the representations from a discriminatively trained Convolutional Neural Network (CNN). 
Our algorithm pretrains a CNN using a large set of videos with tracking ground-truths to obtain a generic target representation. 
Our network is composed of shared layers and multiple branches of domain-specific layers, where domains correspond to individual training sequences and each branch is responsible for binary classification to identify the target in each domain.
We train the network with respect to each domain iteratively to obtain generic target representations in the shared layers. 
When tracking a target in a new sequence, we construct a new network by combining the shared layers in the pretrained CNN with a new binary classification layer, which is updated online.
Online tracking is performed by evaluating the candidate windows randomly sampled around the previous target state. 
The proposed algorithm illustrates outstanding performance compared with state-of-the-art methods in existing tracking benchmarks.
\end{abstract}

\section{Introduction}

Convolutional Neural Networks (CNNs) have recently been applied to various computer vision tasks such as image classification~\cite{krizhevsky2012imagenet,chatfield2014return,SimonyanICLR15}, semantic segmentation~\cite{long2014fully}, object detection~\cite{girshick2014rich}, and many others~\cite{toshev2014deeppose,taigman2014deepface}.
Such great success of CNNs is mostly attributed to their outstanding performance in representing visual data.
Visual tracking, however, has been less affected by these popular trends since it is difficult to collect a large amount of training data for video processing applications and training algorithms specialized for visual tracking are not available yet, while the approaches based on low-level handcraft features still work well in practice~\cite{henriques2015high,danelljan2014accurate,hong2015multi,zhang2014meem}. 
Several recent tracking algorithms~\cite{hong2015online,wang2015transferring} have addressed the data deficiency issue by transferring pretrained CNNs on a large-scale classification dataset such as ImageNet~\cite{ILSVRC15}. 
Although these methods may be sufficient to obtain generic feature representations, its effectiveness in terms of tracking is limited due to the fundamental inconsistency between classification and tracking problems, \ie, predicting object class labels versus locating targets of arbitrary classes.

To fully exploit the representation power of CNNs in visual tracking, it is desirable to train them on large-scale data specialized for visual tracking, which cover a wide range of variations in the combination of target and background.
However, it is truly challenging to learn a unified representation based on the video sequences that have completely different characteristics.
Note that individual sequences involve different types of targets whose class labels, moving patterns, and appearances are different, and tracking algorithms suffer from sequence-specific challenges including occlusion, deformation, lighting condition change, motion blur, etc.
Training CNNs is even more difficult since the same kind of objects can be considered as a target in a sequence and as a background object in another.
Due to such variations and inconsistencies across sequences, we believe that the ordinary learning methods based on the standard classification task are not appropriate, and another approach to capture sequence-independent information should be incorporated for better representations for tracking.

Motivated by this fact, we propose a novel CNN architecture, referred to as {\em Multi-Domain Network} (MDNet), to learn the shared representation of targets from multiple annotated video sequences for tracking, where each video is regarded as a separate domain.
The proposed network has separate branches of domain-specific layers for binary classification at the end of the network, and shares the common information captured from all sequences in the preceding layers for generic representation learning.
Each domain in MDNet is trained separately and iteratively while the shared layers are updated in every iteration.
By employing this strategy, we separate domain-independent information from domain-specific one and learn generic feature representations for visual tracking.
Another interesting aspect of our architecture is that we design the CNN with a small number of layers compared to the networks for classification tasks such as AlexNet~\cite{krizhevsky2012imagenet} and VGG nets~\cite{chatfield2014return,SimonyanICLR15}.

We also propose an effective online tracking framework based on the representations learned by MDNet. 
When a test sequence is given, all the existing branches of binary classification layers, which were used in the training phase, are removed and a new single branch is constructed to compute the target scores in the test sequence. 
The new classification layer and the fully connected layers within the shared layers are then fine-tuned online during tracking to adapt to the new domain.
The online update is conducted to model long-term and short-term appearance variations of a target for robustness and adaptiveness, respectively, and an effective and efficient hard negative mining technique is incorporated in the learning procedure. 

Our algorithm consists of multi-domain representation learning and online visual tracking.
The main contributions of our work are summarized below:
\begin{itemize}
\item We propose a multi-domain learning framework based on CNNs, which separates domain-independent information from domain-specific one, to capture shared representations effectively.
\item Our framework is successfully applied to visual tracking, where the CNN pretrained by multi-domain learning is updated online in the context of a new sequence to learn domain-specific information adaptively.
\item Our extensive experiment demonstrates the outstanding performance of our tracking algorithm compared to the state-of-the-art techniques in two public benchmarks: Object Tracking Benchmark~\cite{otb2} and VOT2014~\cite{vot14}.
\end{itemize}

The rest of the paper is organized as follows.
We first review related work in Section~\ref{sec:related}, and discuss our multi-domain learning approach for visual tracking in Section~\ref{sec:multi-domain}.
Section~\ref{sec:online} describes our online learning and tracking algorithm, and Section~\ref{sec:experiment} demonstrates the experimental results in two tracking benchmark datasets.

\section{Related Work}
\label{sec:related}

\subsection{Visual Tracking Algorithms}
Visual tracking is one of the fundamental problems in computer vision and has been actively studied for decades. 
Most tracking algorithms fall into either generative or discriminative approaches. 
Generative methods describe the target appearances using generative models and search for the target regions that fit the models best. 
Various generative target appearance modeling algorithms have been proposed including sparse representation~\cite{mei2009robust,zhang2012robust}, density estimation~\cite{HanTPAMI08,JepsonTPAMI03}, and incremental subspace learning~\cite{ross2008incremental}. 
In contrast, discriminate methods aim to build a model that distinguishes the target object from the background.  
These tracking algorithms typically learn classifiers based on multiple instance learning~\cite{BabenkoTPAMI11}, P-N learning~\cite{kalal2012tracking}, online boosting~\cite{GrabnerBMVC06,GrabnerECCV08}, structured output SVMs~\cite{hare2011struck}, etc.

In recent years, correlation filters have gained attention in the area of visual tracking due to their computational efficiency and competitive performance~\cite{BolmeCVPR10,henriques2015high,danelljan2014accurate,hong2015multi}. 
Bolme \etal~\cite{BolmeCVPR10} proposed a fast correlation tracker with a minimum output sum of squared error (MOSSE) filter, which runs in hundreds of frames per second. 
Henriques \etal~\cite{henriques2015high} formulated  kernelized correlation filters (KCF) using circulant matrices, and efficiently incorporated multi-channel features in a Fourier domain. 
Several variations of KCF tracker have been subsequently investigated to improve tracking performance.
For example, DSST~\cite{danelljan2014accurate} learns separate filters for translation and scaling, and MUSTer~\cite{hong2015multi} employs short-term and long-term memory stores inspired by a psychological memory model. 
Although these approaches are satisfactory in constrained environments, they have an inherent limitation that they resort to low-level hand-crafted features, which are vulnerable in dynamic situations including illumination changes, occlusion, deformations, etc.  

\begin{figure*}
\begin{center}
\includegraphics[width=0.68\linewidth]{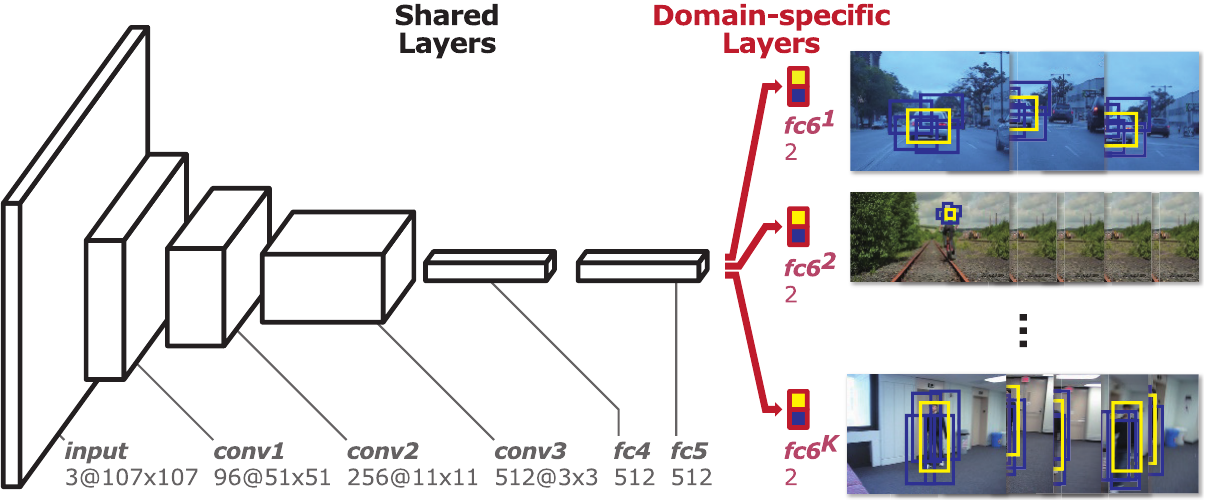}
\end{center}
\vspace{-6mm}
   \caption{The architecture of our Multi-Domain Network, which consists of shared layers and $K$ branches of domain-specific layers. Yellow and blue bounding boxes denote the positive and negative samples in each domain, respectively.}
\label{fig:arch}
\end{figure*}

\subsection{Convolutional Neural Networks}
CNNs have demonstrated their outstanding representation power in a wide range of computer vision applications~\cite{krizhevsky2012imagenet,chatfield2014return,SimonyanICLR15,girshick2014rich,long2014fully,toshev2014deeppose,taigman2014deepface}.  
Krizhevsky \etal~\cite{krizhevsky2012imagenet} brought significant performance improvement in image classification by training a deep CNN with a large-scale dataset and an efficient GPU implementation. 
R-CNN~\cite{girshick2014rich} applies a CNN to an object detection task, where the training data are scarce, by pretraining on a large auxiliary dataset and fine-tuning on the target dataset. 

Despite such huge success of CNNs, only a limited number of tracking algorithms using the representations from CNNs have been proposed so far~\cite{fan2010human,hong2015online,li122014deeptrack,wang2015transferring}.
An early tracking algorithm based on a CNN can handle only predefined target object classes, \eg, human, since the CNN is trained offline before tracking and fixed afterwards~\cite{fan2010human}.
Although \cite{li122014deeptrack} proposes an online learning method based on a pool of CNNs, it suffers from lack of training data to train deep networks and its accuracy is not particularly good compared to the methods based on hand-craft features.
A few recent approaches~\cite{wang2015transferring,hong2015online} transfer CNNs pretrained on a large-scale dataset constructed for image classification, but the representation may not be very effective due to the fundamental difference between classification and tracking tasks. 
Contrary to the existing approaches, our algorithm takes advantage of large-scale visual tracking data for pretraining a CNN and obtain effective representations.

\subsection{Multi-Domain Learning}
\label{sub:multi-domain}
Our approach to pretrain deep CNNs belongs to multi-domain learning, which refers to a learning method in which the training data are originated from multiple domains and the domain information is incorporated in learning procedure. 
Multi-domain learning is popular in natural language processing (\eg, sentiment classification with multiple products and spam filtering with multiple users), and various approaches have been proposed~\cite{daume2009frustratingly,dredze2010multi,joshi2012multi}. 
In computer vision community, multi-domain learning is discussed in only a few domain adaptation approaches. 
For example, Duan \etal~\cite{duan2009domain} introduced a domain-weighted combination of SVMs for video concept detection, and Hoffman \etal~\cite{hoffman2012discovering} presented a mixture-transform model for object classification.

\section{Multi-Domain Network (MDNet)}
\label{sec:multi-domain}
This section describes our CNN architecture and multi-domain learning approach to obtain domain-independent representations for visual tracking.

\subsection{Network Architecture}
The architecture of our network is illustrated in Figure~\ref{fig:arch}.
It receives a 107$\times$107 RGB input\footnote{This input size is designed to obtain 3$\times$3 feature maps in conv3: $107= 75~(\text{receptive field}) + 2\times16~(\text{stride})$.}, and has five hidden layers including three convolutional layers (conv1-3) and two fully connected layers (fc4-5).
Additionally, the network has $K$ branches for the last fully connected layers (fc6\textsuperscript{1}-fc6\textsuperscript{$K$}) corresponding to $K$ domains, in other words, training sequences.
The convolutional layers are identical to the corresponding parts of VGG-M network~\cite{chatfield2014return} except that the feature map sizes are adjusted by our input size.
The next two fully connected layers have 512 output units and are combined with ReLUs and dropouts.
Each of the $K$ branches contains a binary classification layer with softmax cross-entropy loss, which is responsible for distinguishing target and background in each domain.
Note that we refer to fc6\textsuperscript{1}-fc6\textsuperscript{$K$} as domain-specific layers and all the preceding layers as shared layers.

Our network architecture is substantially smaller than the ones commonly used in typical recognition tasks such as AlexNet~\cite{krizhevsky2012imagenet} and VGG-Nets~\cite{chatfield2014return,SimonyanICLR15}.
We believe that such a simple architecture is more appropriate for visual tracking due to the following reasons. 
First, visual tracking aims to distinguish only two classes, target and background, which requires much less model complexity than general visual recognition problems such as ImageNet classification with 1000 classes.
Second, a deep CNN is less effective for precise target localization since the spatial information tends to be diluted as a network goes deeper~\cite{hong2015online}.
Third, since targets in visual tracking are typically small, it is desirable to make input size small, which reduces the depth of the network naturally.
Finally, a smaller network is obviously more efficient in visual tracking problem, where training and testing are performed online.
When we tested larger networks, the algorithm is less accurate and becomes slower significantly.

\subsection{Learning Algorithm}
\label{sec:mdl}
The goal of our learning algorithm is to train a multi-domain CNN disambiguating target and background in an arbitrary domain, which is not straightforward since the training data from different domains have different notions of target and background.
However, there still exist some common properties that are desirable for target representations in all domains, such as robustness to illumination changes, motion blur, scale variations, etc.
To extract useful features satisfying these common properties, we separate domain-independent  information from domain-specific one by incorporating a multi-domain learning framework.

Our CNN is trained by the Stochastic Gradient Descent (SGD) method, where each domain is handled exclusively in each iteration. 
In the $k^{\text{th}}$ iteration, the network is updated based on a minibatch that consists of the training samples from the $(k \operatorname{mod} K)^{\text{th}}$ sequence, where only a single branch fc6\textsuperscript{$(k \operatorname{mod} K)$} is enabled. 
It is repeated until the network is converged or the predefined number of iterations is reached.
Through this learning procedure, domain-independent information is modeled in the shared layers from which useful generic feature representations are obtained.

\section{Online Tracking using MDNet}
\label{sec:online}
Once we complete the multi-domain learning described in Section~\ref{sec:mdl}, the multiple branches of domain-specific layers (fc6\textsuperscript{1}-fc6\textsuperscript{$K$}) are replaced with a single branch (fc6) for a new test sequence. 
Then we fine-tune the new domain-specific layer and the fully connected layers in the shared network online at the same time.
The detailed tracking procedure is discussed in this section.

\subsection{Tracking Control and Network Update}
We consider two complementary aspects in visual tracking, robustness and adaptiveness, by long-term and short-term updates.
Long-term updates are performed in regular intervals using the positive samples collected for a long period of time while short-term updates are conducted whenever potential tracking failures are detected---when the estimated target is classified as background---using the positive samples in a short-term period.
In both cases we use the negative samples observed in the short-term since old negative examples are often redundant or irrelevant to the current frame.
Note that we maintain a single network during tracking, where these two kinds of updates are performed depending on how fast the target appearance changes.

To estimate the target state in each frame, $N$ target candidates $\mathbf{x}^1,\dots,\mathbf{x}^N$ sampled around the previous target state are evaluated using the network, and we obtain their positive scores $f^+(\mathbf{x}^i)$ and negative scores $f^-(\mathbf{x}^i)$ from the network. 
The optimal target state $\mathbf{x}^*$ is given by finding the example with the maximum positive score as
\begin{equation}
\label{eq:target}
\mathbf{x}^* = \arg\!\max_{\mathbf{x}^i} f^+(\mathbf{x}^i).
\end{equation}

\subsection{Hard Minibatch Mining}

\begin{figure}[t]
\begin{center}
\includegraphics[width=0.32\linewidth]{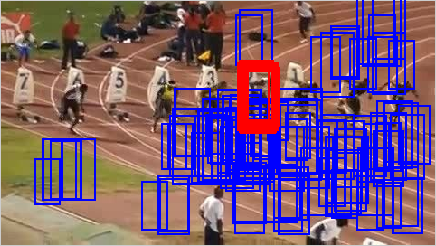}
\includegraphics[width=0.32\linewidth]{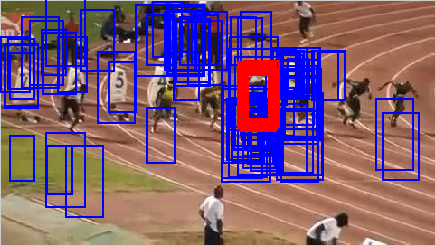}
\includegraphics[width=0.32\linewidth]{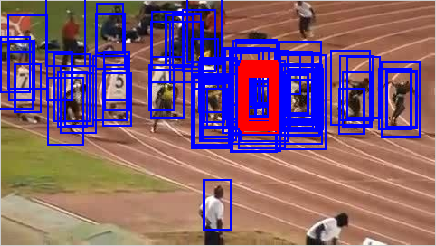}

\begin{subfigure}[b]{0.32\linewidth}
\includegraphics[width=\linewidth]{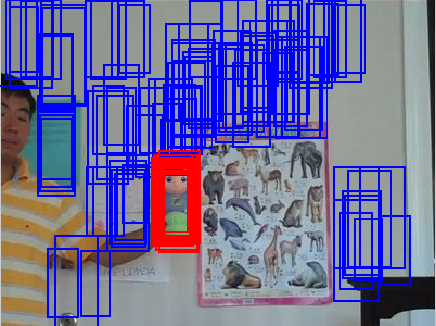}
\vspace{-5mm}
\caption{$1^{\rm st}$ minibatch}
\end{subfigure}
\begin{subfigure}[b]{0.32\linewidth}
\includegraphics[width=\linewidth]{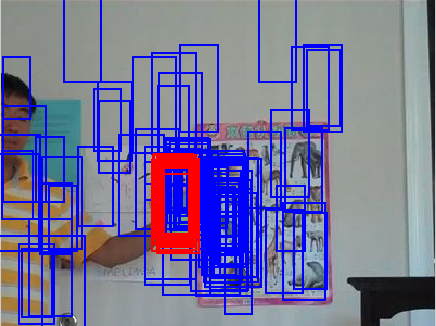}
\vspace{-5mm}
\caption{$5^{\rm th}$ minibatch}
\end{subfigure}
\begin{subfigure}[b]{0.32\linewidth}
\includegraphics[width=\linewidth]{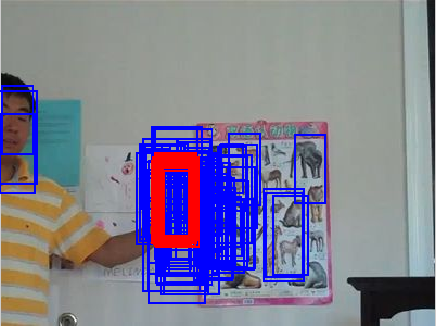}
\vspace{-5mm}
\caption{$30^{\rm th}$ minibatch}
\end{subfigure}
\end{center}
\vspace{-5mm}
\caption{Identified training examples through our hard negative mining in \emph{Bolt2} (top) and \emph{Doll} (bottom) sequences. Red and blue bounding boxes denote positive and negative samples in each minibatch, respectively. The negative samples becomes hard to classify as training proceeds.}
\label{fig:hbm}
\end{figure}

The majority of negative examples are typically trivial or redundant in tracking-by-detection approaches, while only a few distracting negative samples are effective in training a classifier. 
Hence, the ordinary SGD method, where the training samples evenly contribute to learning, easily suffers from a drift problem since the distractors are considered insufficiently. 
A popular solution in object detection for this issue is hard negative mining~\cite{655648}, where training and testing procedures are alternated to identify the hard negative examples, typically false positives, and we adopt this idea for our online learning procedure.

We integrate hard negative mining step into minibatch selection. 
In each iteration of our learning procedure, a minibatch consists of $M^+$ positives and $M_h^-$ hard negatives. 
The hard negative examples are identified by testing $M^- (\gg M_h^-)$ negative samples and selecting the ones with top $M_h^-$ positive scores. 
As the learning proceeds and the network becomes more discriminative, the classification in a minibatch becomes more challenging as illustrated in Figure~\ref{fig:hbm}. 
This approach examines a predefined number of samples and identifies critical negative examples effectively without explicitly running a detector to extract false positives as in the standard hard negative mining techniques.

\subsection{Bounding Box Regression}
Due to the high-level abstraction of CNN-based features and our data augmentation strategy which samples multiple positive examples around the target (which will be described in more detail in the next subsection), our network sometimes fails to find tight bounding boxes enclosing the target.
We apply the bounding box regression technique, which is popular in object detection~\cite{girshick2014rich,felzenszwalb2010object}, to improve target localization accuracy.
Given the first frame of a test sequence, we train a simple linear regression model to predict the precise target location using conv3 features of the samples near the target location.
In the subsequent frames, we adjust the target locations estimated from Eq.~\eqref{eq:target} using the regression model if the estimated targets are reliable (\ie $f^+(x^*)>0.5$).
The bounding box regressor is trained only in the first frame since it is time consuming for online update and incremental learning of the regression model may not be very helpful considering its risk.
Refer to \cite{girshick2014rich} for details as we use the same formulation and parameters.

\subsection{Implementation Details}
\label{sub:implementation}

\begin{algorithm}[t]
\caption{Online tracking algorithm}
\label{alg:tracking}
\begin{algorithmic}[1]
\Require Pretrained CNN filters $\left\{ \mathbf{w}_{1}, \dots, \mathbf{w}_{5}\right\}$
  \Statex ~~~~~~~~~Initial target state $\mathbf{x}_1$
\Ensure Estimated target states $\mathbf{x}_t^*$
\State Randomly initialize the last layer $\mathbf{w}_6$.
\State Train a bounding box regression model.
\State Draw positive samples $S_1^+$ and negative samples $S_1^-$.
\State Update $\left\{\mathbf{w}_4,\mathbf{w}_5,\mathbf{w}_6\right\}$ using $S_1^+$ and $S_1^-$;
\State $\mathcal{T}_s \gets \{1\}$ and $\mathcal{T}_l \gets \{1\}$.

\Repeat
    \State Draw target candidate samples $\mathbf{x}_t^i$.
    \State Find the optimal target state $\mathbf{x}_t^*$ by Eq.~\eqref{eq:target}.
    \If {$f^+(\mathbf{x}_t^*) > 0.5$}
        \State Draw training samples $S_t^+$ and $S_t^-$.
        \State $\mathcal{T}_s \gets \mathcal{T}_s \cup \{t\}$, $\mathcal{T}_l \gets \mathcal{T}_l \cup \{t\}$.
        \If {$|\mathcal{T}_s|>\tau_s$} $\mathcal{T}_s \gets \mathcal{T}_s \setminus \{\min_{v\in \mathcal{T}_s} v$\}. \EndIf
        \If {$|\mathcal{T}_l|>\tau_l$} $\mathcal{T}_l \gets \mathcal{T}_l \setminus \{\min_{v\in \mathcal{T}_l} v$\}. \EndIf
        \State Adjust $\mathbf{x}_t^*$ using bounding box regression.
    \EndIf
\If {$f^+(\mathbf{x}_t^*)< 0.5$}
    \State Update $\left\{\mathbf{w}_4,\mathbf{w}_5,\mathbf{w}_6\right\}$ using $S^+_{v\in\mathcal{T}_s}$ and $S^-_{v\in\mathcal{T}_s}$.
\ElsIf{$t \operatorname{mod} 10 = 0$}
    \State Update $\left\{\mathbf{w}_4,\mathbf{w}_5,\mathbf{w}_6\right\}$ using $S^+_{v\in\mathcal{T}_l}$ and $S^-_{v\in\mathcal{T}_s}$.
\EndIf  
\Until {end of sequence}
\end{algorithmic}
\end{algorithm}

The overall procedure of our tracking algorithm is presented in Algorithm~\ref{alg:tracking}. 
The filter weights in the $j^{\text{th}}$ layer of CNN are denoted by $\mathbf{w}_j$, where $\mathbf{w}_{1:5}$ are pretrained by mutli-domain learning and $\mathbf{w}_6$ is initialized randomly for a new sequence.
Only the weights in the fully connected layers $\mathbf{w}_{4:6}$ are updated online whereas the ones in the convolutional layers $\mathbf{w}_{1:3}$ are fixed throughout tracking; this strategy is beneficial to not only  computational efficiency but also avoiding overfitting by preserving domain-independent information. 
$\mathcal{T}_s$ and $\mathcal{T}_l$ are frame index sets in short-term ($\tau_s=20$) and long-term ($\tau_l=100$) periods, respectively.
The further implementation details are described below.

\paragraph{Target candidate generation}
To generate target candidates in each frame, we draw $N (=256)$ samples in translation and scale dimension, $\mathbf{x}_t^i=(x_t^i, y_t^i, s_t^i)$, $i=1,\dots,N$, from a Gaussian distribution whose mean is the previous target state $\mathbf{x}_{t-1}^*$ and covariance is a diagonal matrix $\text{diag}(0.09r^2, 0.09r^2, 0.25)$, where $r$ is the mean of the width and height of the target in the previous frame. 
The scale of each candidate bounding box is computed by multiplying $1.05^{s_i}$ to the initial target scale.

\paragraph{Training data}
For offline multi-domain learning, we collect 50 positive and 200 negative samples from every frame, where positive and negative examples have $\geq 0.7$ and $\leq 0.5$ IoU overlap ratios with ground-truth bounding boxes, respectively. 
Similarly, for online learning, we collect $S_t^+ (=50)$ positive and $S_t^- (=200)$ negative samples with $\geq 0.7$ and $\leq 0.3$ IoU overlap ratios with the estimated target bounding boxes, respectively, except that $S_1^+=500$ and $S_1^-=5000$.
For bounding-box regression, we use 1000 training examples with the same parameters as \cite{girshick2014rich}.

\paragraph{Network learning}
For multi-domain learning with $K$ training sequences, we train the network for $100K$ iterations with learning rates 0.0001 for convolutional layers\footnote{The convolutional layers are initialized by VGG-M network, which is pretrained on ImageNet.} and 0.001 for fully connected layers.
At the initial frame of a test sequence, we train the fully connected layers for 30 iterations with learning rate 0.0001 for fc4-5 and 0.001 for fc6.
For online update, we train the fully connected layers for 10 iterations with the learning rate three times larger than that in the initial frame for fast adaptation.
The momentum and weight decay are always set to 0.9 and 0.0005, respectively.
Each mini-batch consists of $M^+ (=32)$ positives and  $M_h^- (=96)$ hard negatives selected out of $M^- (=1024)$ negative examples.

\section{Experiment}
\label{sec:experiment}
We evaluated the proposed tracking algorithm on two public benchmark datasets, Object Tracking Benchmark (OTB)~\cite{otb2} and VOT2014~\cite{vot14}, and compared its performance with state-of-the-art trackers. 
Our algorithm is implemented in MATLAB using MatConvNet toolbox~\cite{vedaldi15matconvnet}, and runs at around 1 fps with eight cores of 2.20GHz Intel Xeon E5-2660 and an NVIDIA Tesla K20m GPU.

\subsection{Evaluation on OTB}

\begin{figure}[t]
\begin{center}
\begin{subfigure}[b]{\linewidth}
\includegraphics[width=0.495\linewidth]{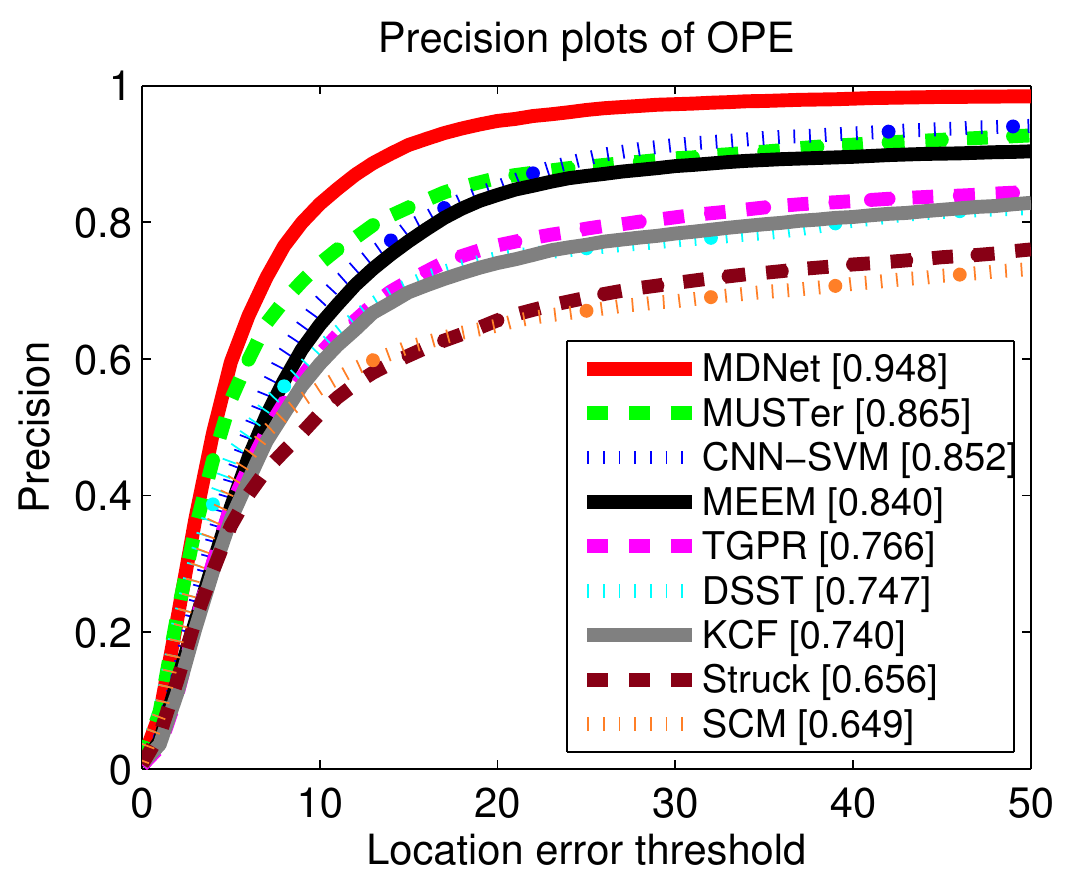}
\includegraphics[width=0.495\linewidth]{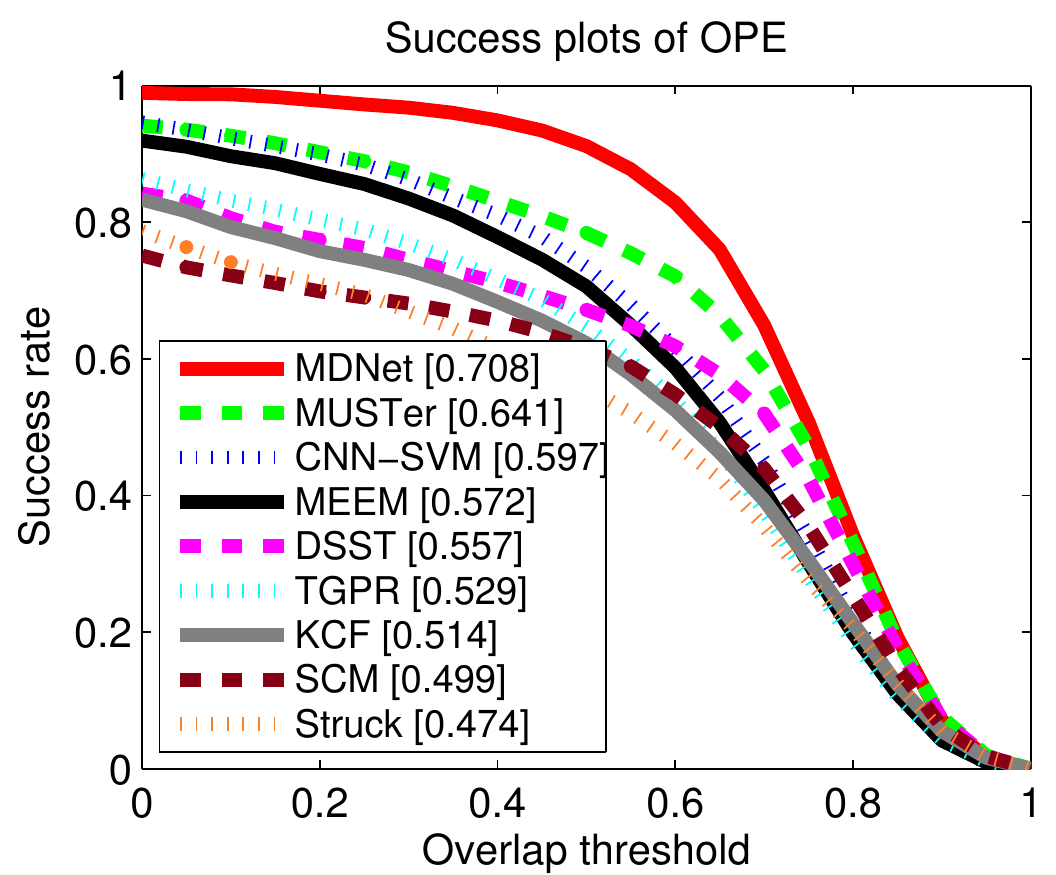}
\vspace{-5mm}
\caption{OTB50 result}
\label{fig:otb50}
\end{subfigure}
\begin{subfigure}[b]{\linewidth}
\includegraphics[width=0.495\linewidth]{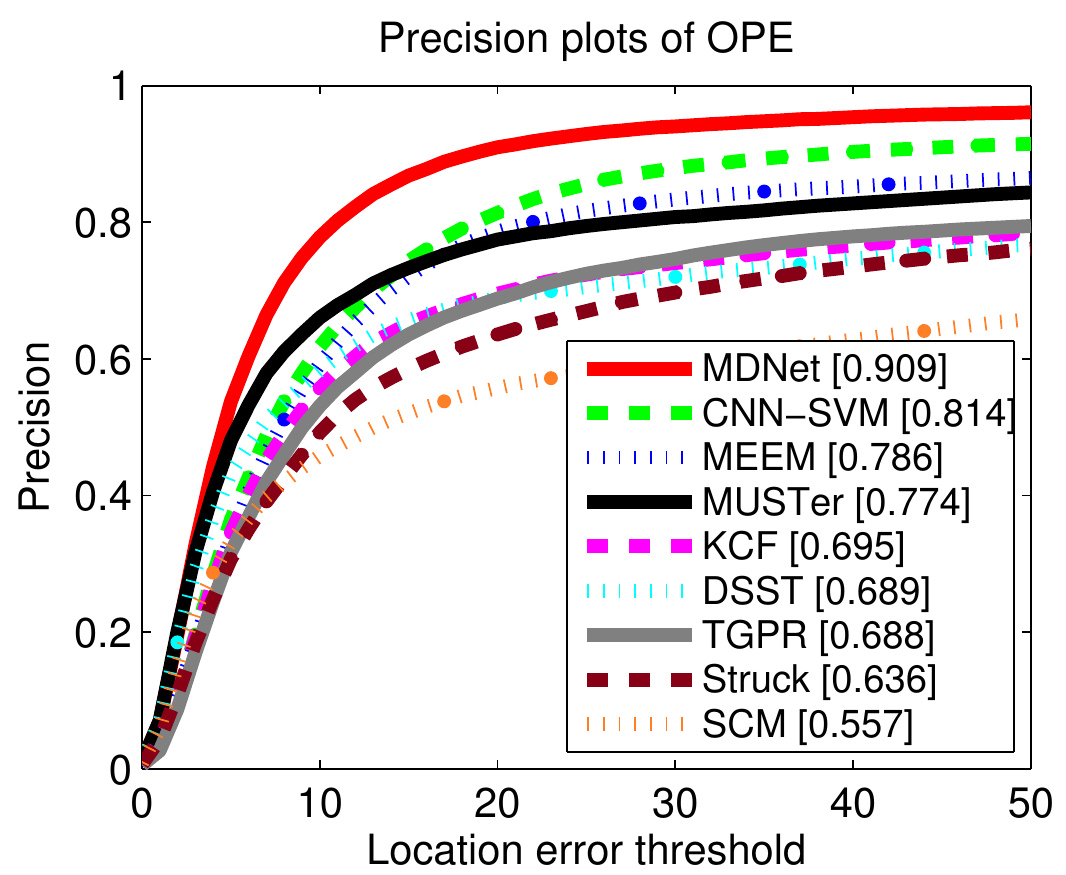}
\includegraphics[width=0.495\linewidth]{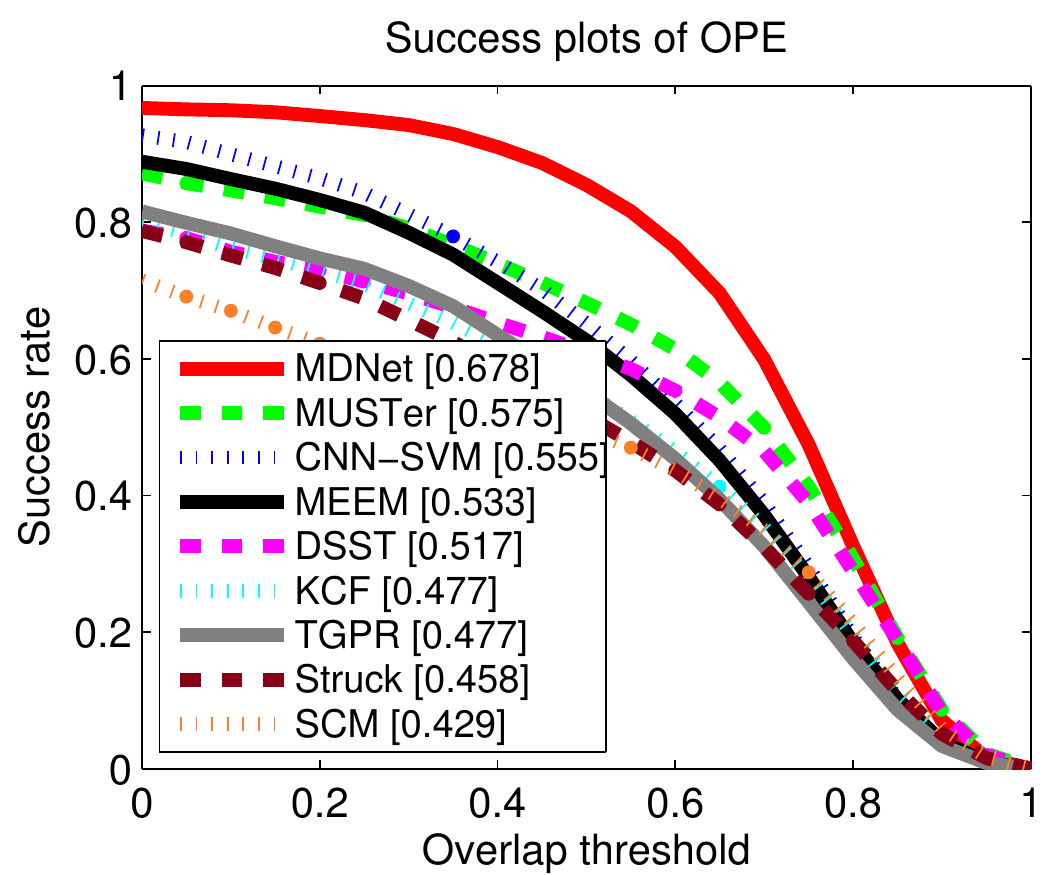}
\vspace{-5mm}
\caption{OTB100 result}
\label{fig:otb100}
\end{subfigure}
\end{center}
\vspace{-5mm}
\caption{Precision and success plots on OTB50~\cite{otb1} and OTB100~\cite{otb2}. The numbers in the legend indicate the representative precisions at 20 pixels for precision plots, and the area-under-curve scores for success plots. }
\label{fig:otb_result}
\end{figure}

\begin{figure*}
\begin{center}
\includegraphics[width=0.24\linewidth]{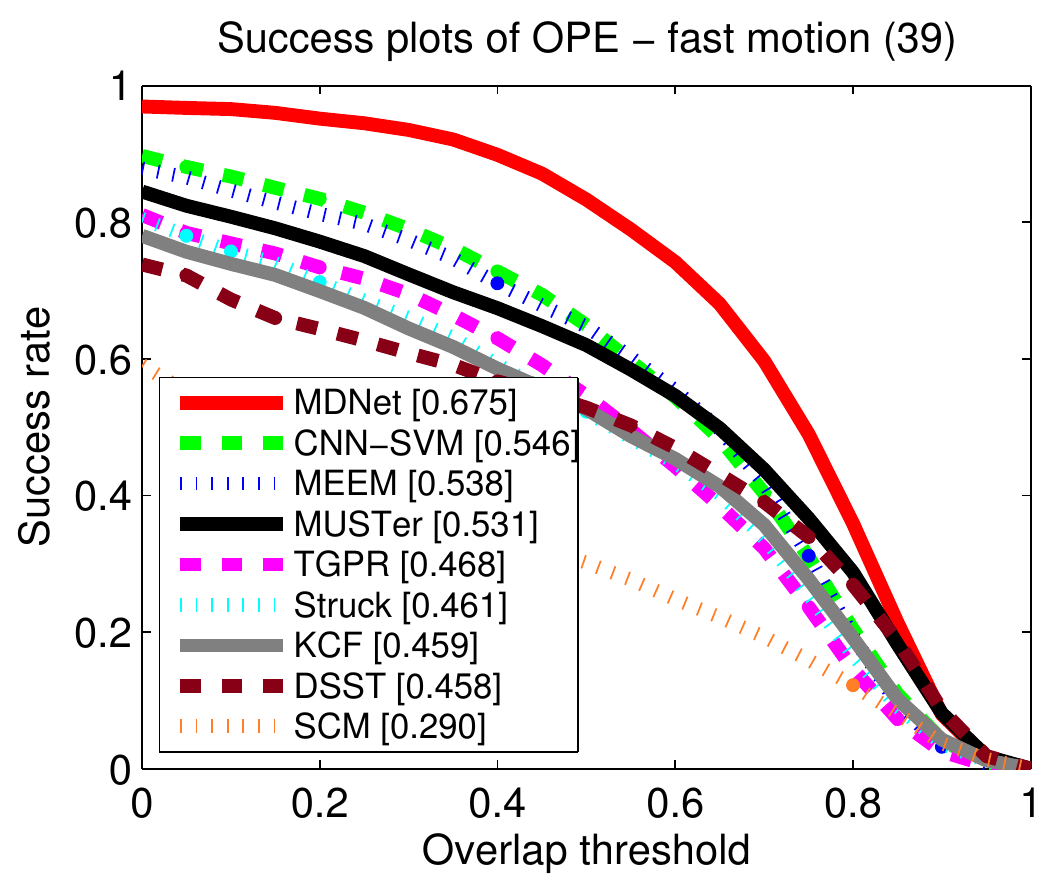}
\includegraphics[width=0.24\linewidth]{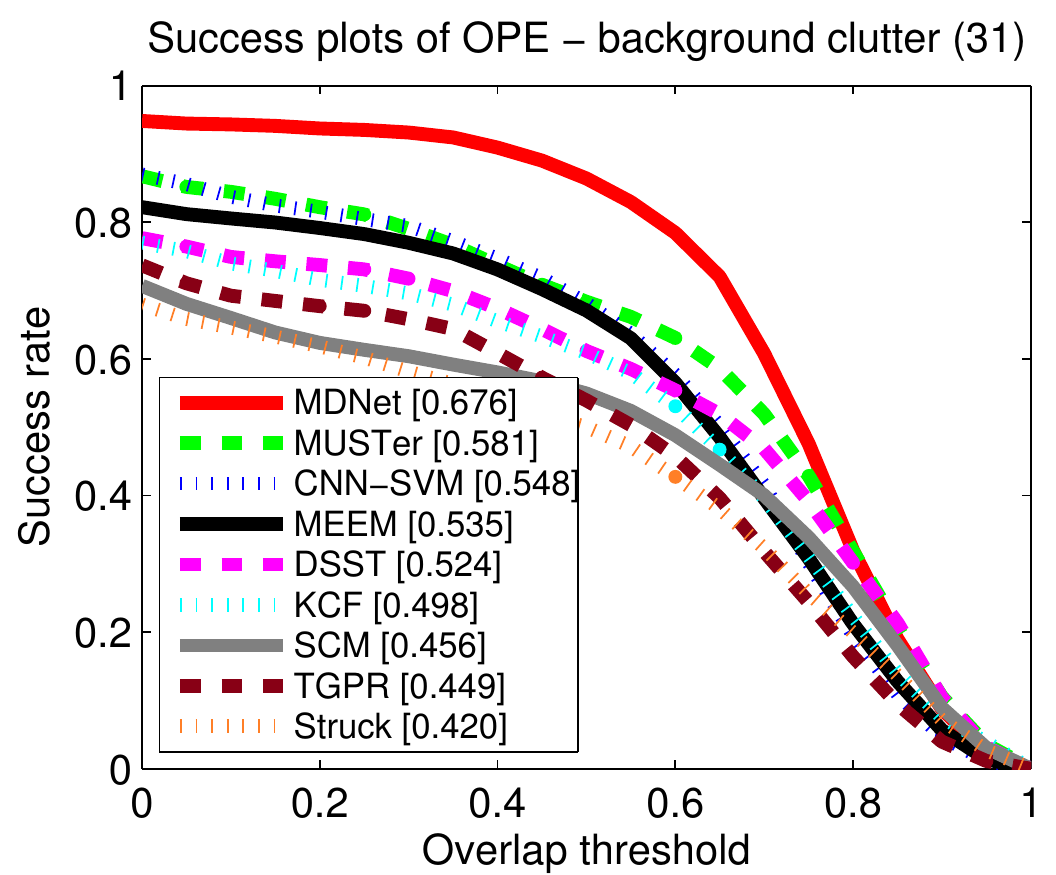}
\includegraphics[width=0.24\linewidth]{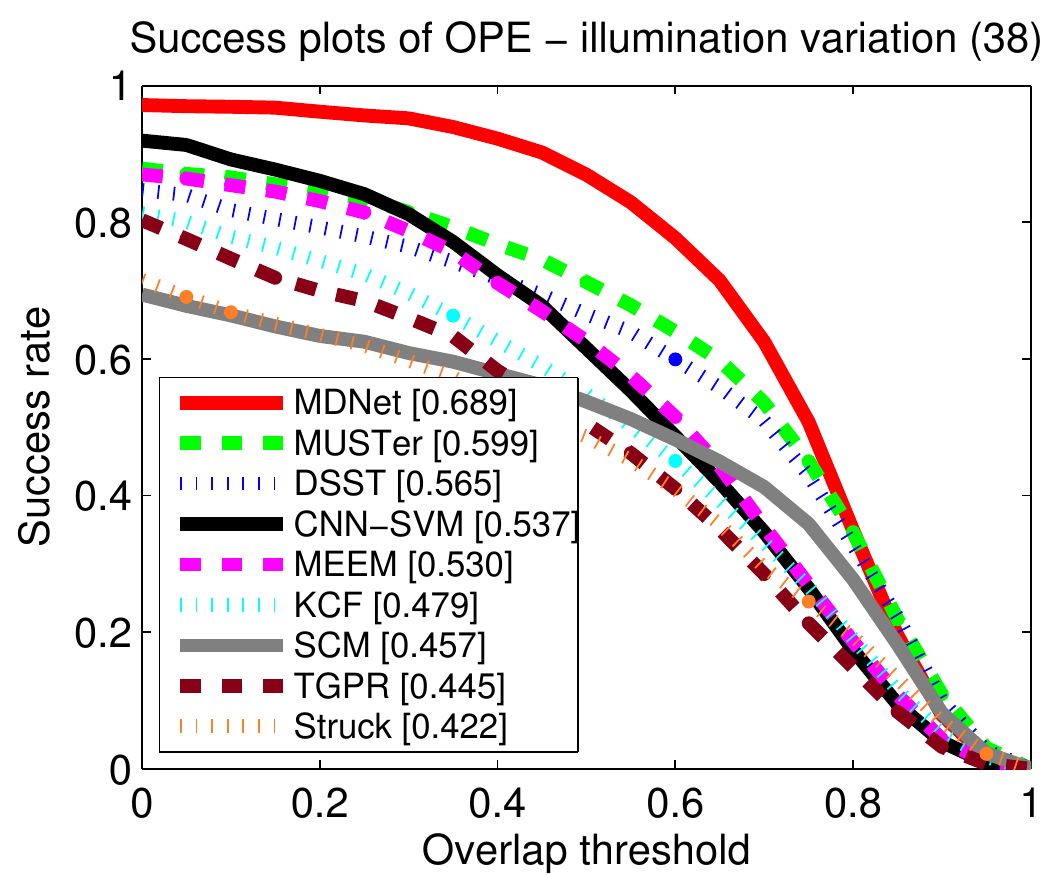}
\includegraphics[width=0.24\linewidth]{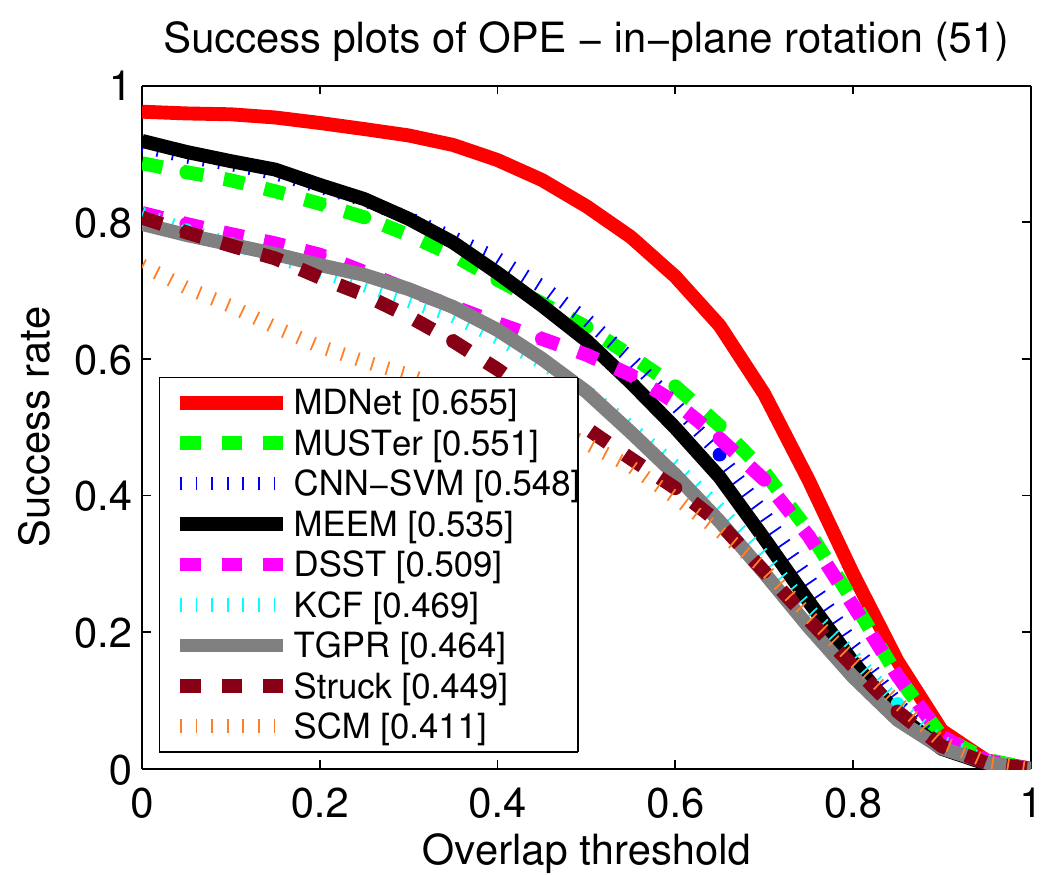}
\includegraphics[width=0.24\linewidth]{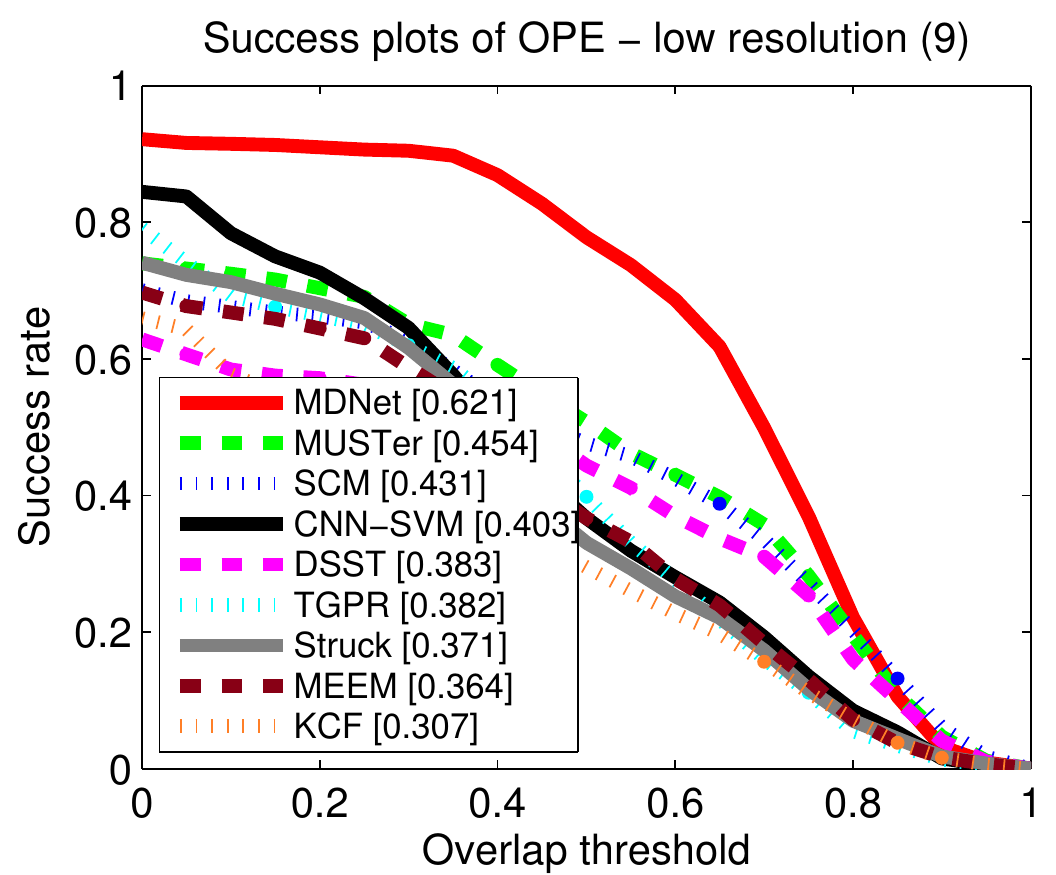}
\includegraphics[width=0.24\linewidth]{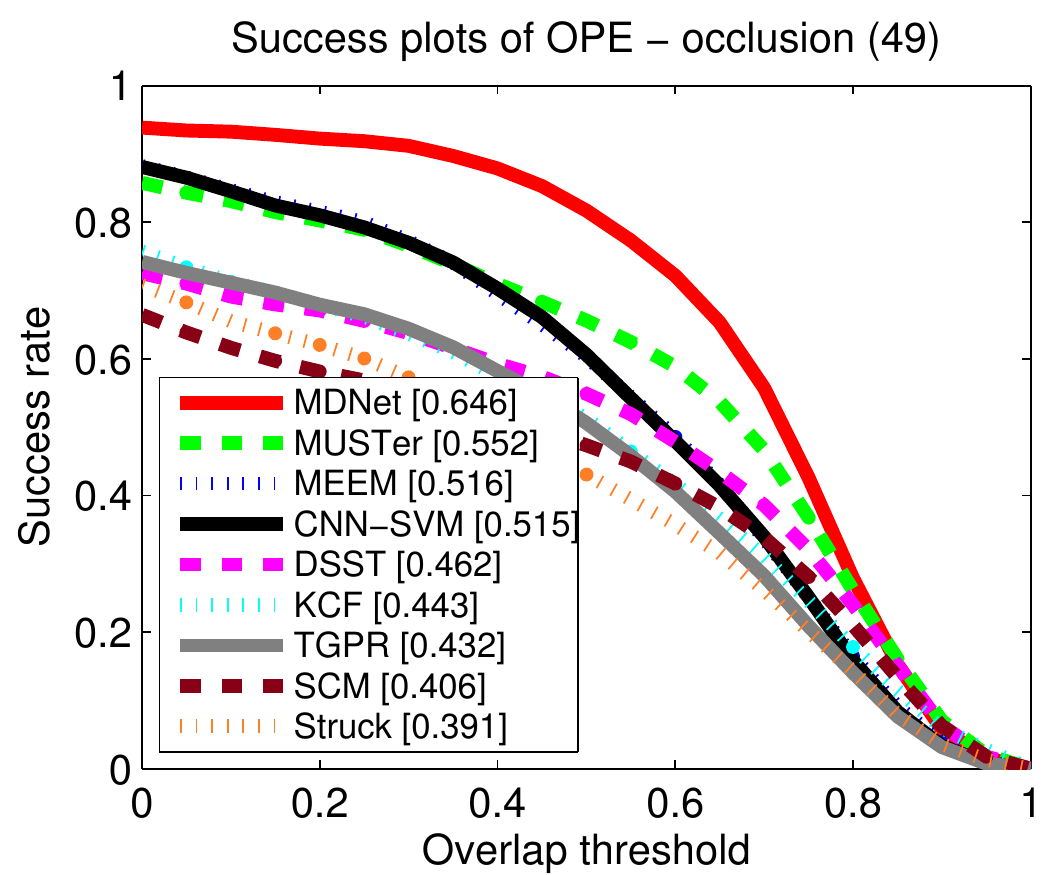}
\includegraphics[width=0.24\linewidth]{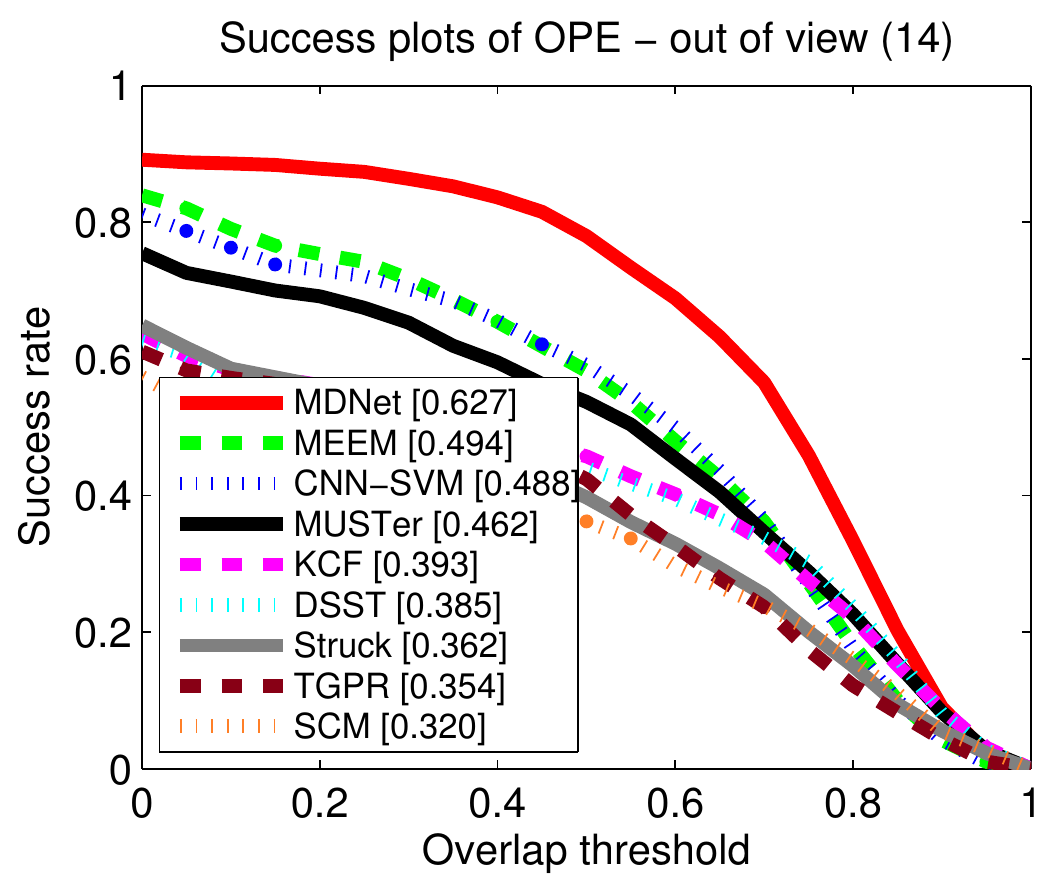}
\includegraphics[width=0.24\linewidth]{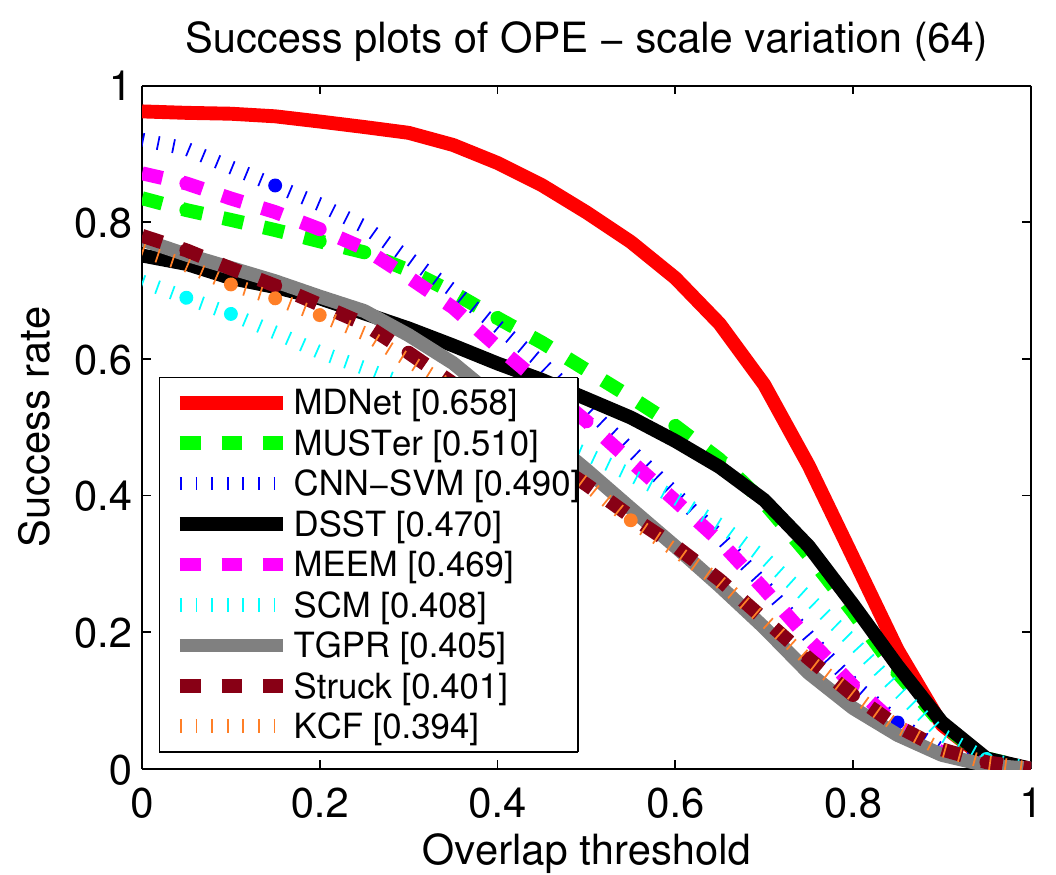}
\end{center}
\vspace{-5mm}
\caption{The success plots for eight challenge attributes: fast motion, background clutter, illumination variation, in-plain rotation, low resolution, occlusion, out of view, and scale variation.}
\label{fig:otb_attributes}
\vspace{-0.5mm}
\end{figure*}

OTB~\cite{otb2} is a popular tracking benchmark that contains 100 fully annotated videos with substantial variations.
The evaluation is based on two metrics: center location error and bounding box overlap ratio.
The one-pass evaluation (OPE) is employed to compare our algorithm with the six state-of-the-art trackers including MUSTer~\cite{hong2015multi}, CNN-SVM~\cite{hong2015online}, MEEM~\cite{zhang2014meem}, TGPR~\cite{gao2014transfer}, DSST~\cite{danelljan2014accurate} and KCF~\cite{henriques2015high}, as well as the top 2 trackers included in the benchmark---SCM~\cite{zhong2012robust} and Struck~\cite{hare2011struck}.
Note that CNN-SVM is another tracking algorithm based on the representations from CNN, which provides a baseline for tracking algorithms that adopt deep learning.
In addition to the results on the entire 100 sequences in \cite{otb2} (OTB100), we also present the results on its earlier version containing 50 sequences~\cite{otb1} (OTB50).
For offline training of MDNet, we use 58 training sequences collected from VOT2013~\cite{vot13}, VOT2014~\cite{vot14} and VOT2015~\cite{vot15}, excluding the videos included in OTB100.

Figure~\ref{fig:otb_result} illustrates the precision and success plots based on center location error and bounding box overlap ratio, respectively.
It clearly illustrates that our algorithm, denoted by MDNet, outperforms the state-of-the-art trackers significantly in both measures.
The exceptional scores at mild thresholds means our tracker hardly misses targets while the competitive scores at strict thresholds implies that our algorithm also finds tight bounding boxes to targets.
For detailed performance analysis, we also report the results on various challenge attributes in OTB100,  such as occlusion, rotation, motion blur, etc.
Figure~\ref{fig:otb_attributes} demonstrates that our tracker effectively handles all kinds of challenging situations that often require high-level semantic understanding.
In particular, our tracker successfully track targets in low resolution while all the trackers based on low-level features are not successful in the challenge.

\begin{figure}[t]
\begin{center}
\includegraphics[width=.495\linewidth]{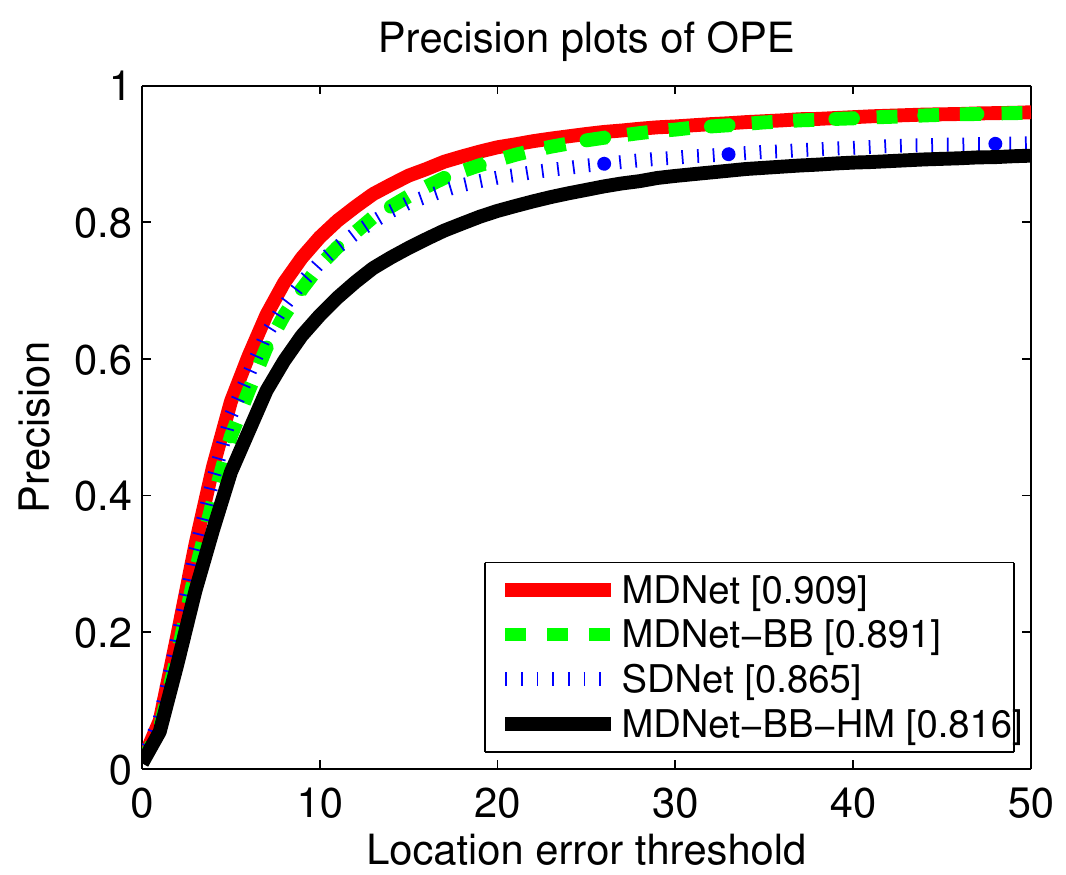}
\includegraphics[width=.495\linewidth]{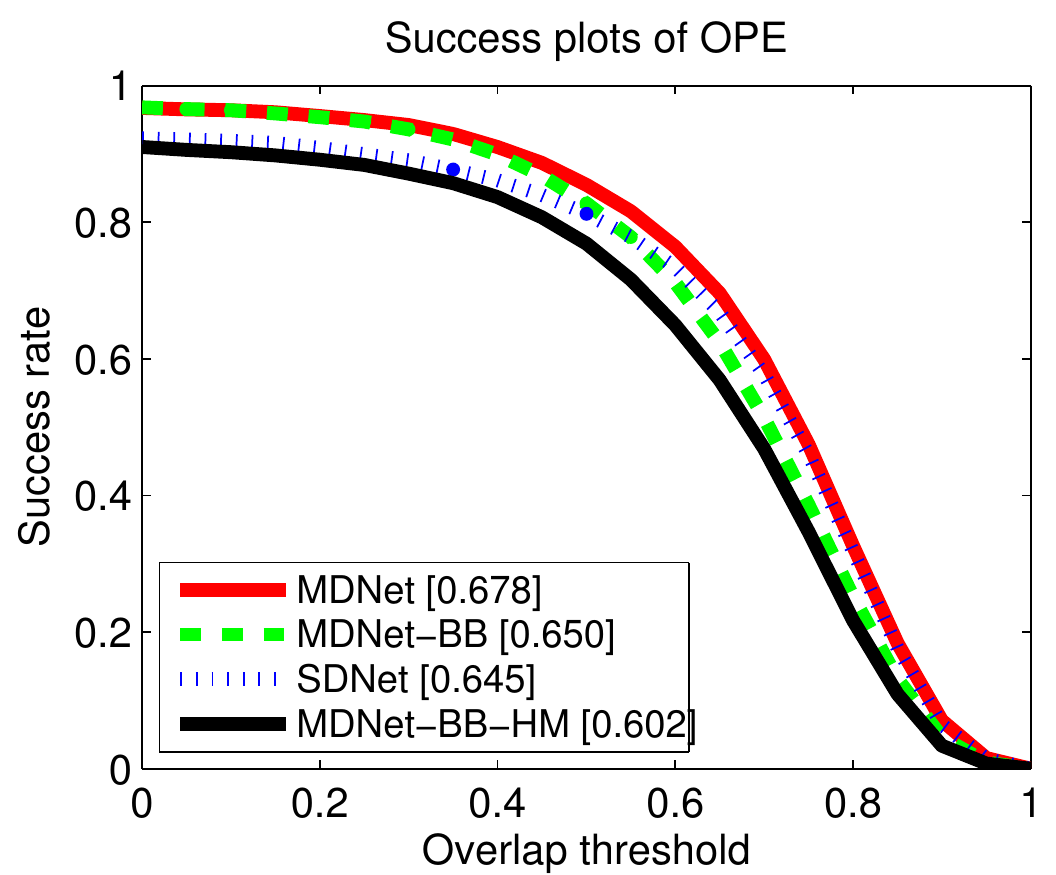}
\end{center}
\vspace{-5mm}
\caption{Precision and success plots on OTB100 for the internal comparisons. }
\label{fig:component}
\vspace{-1mm}
\end{figure}

To verify the contribution of each component in our algorithm, we implement and evaluate several variations of our approach.
The effectiveness of our multi-domain pretraining technique is tested by comparison with the single-domain learning method (SDNet), where the network is trained with a single branch using the data from multiple sequences.
We also investigate two additional versions of our tracking algorithm---MDNet without bounding box regression (MDNet--BB) and MDNet without bounding box regression and hard negative mining (MDNet--BB--HM). 
The performances of all the variations are not as good as our full algorithm (MDNet) and each component in our tracking algorithm is helpful to improve performance.
The detailed results are illustrated in Figure~\ref{fig:component}.

\begin{figure*}
\begin{center}
\includegraphics[width=0.19\linewidth]{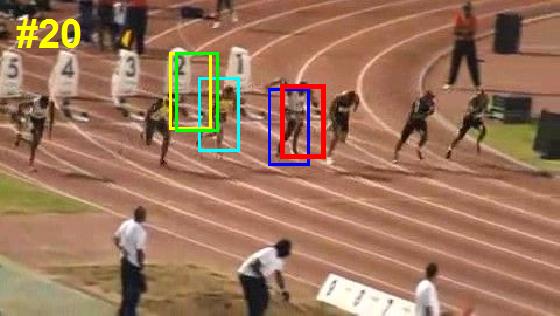}
\includegraphics[width=0.19\linewidth]{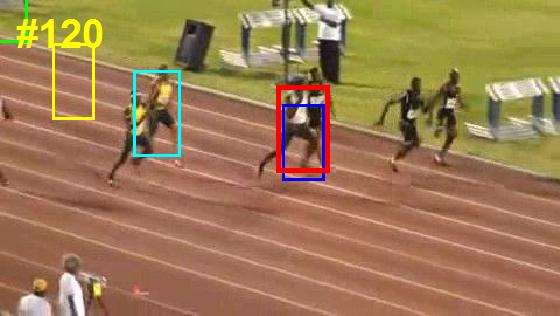}
\includegraphics[width=0.19\linewidth]{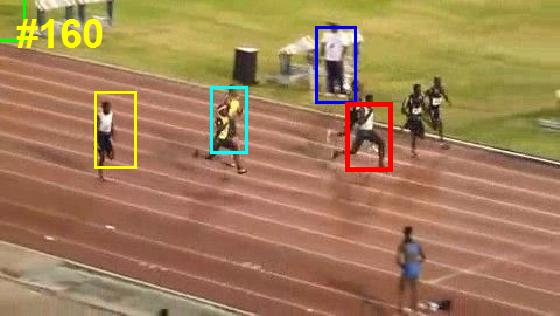}
\includegraphics[width=0.19\linewidth]{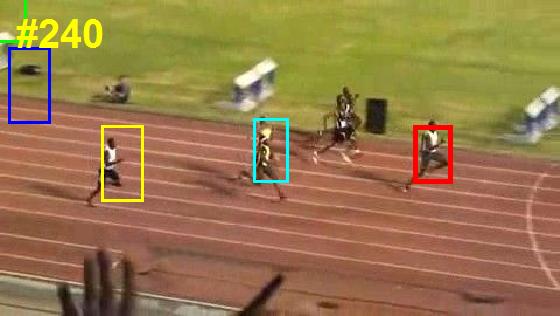}
\includegraphics[width=0.19\linewidth]{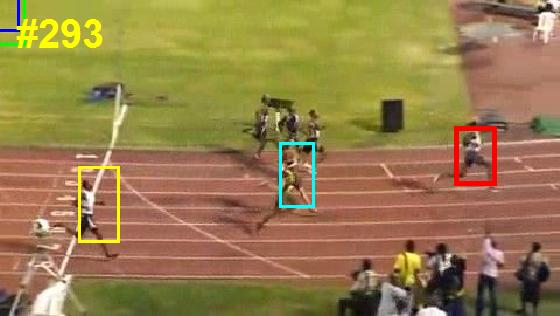}

\includegraphics[width=0.19\linewidth]{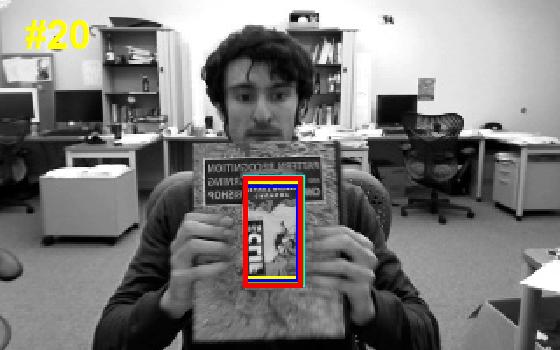}
\includegraphics[width=0.19\linewidth]{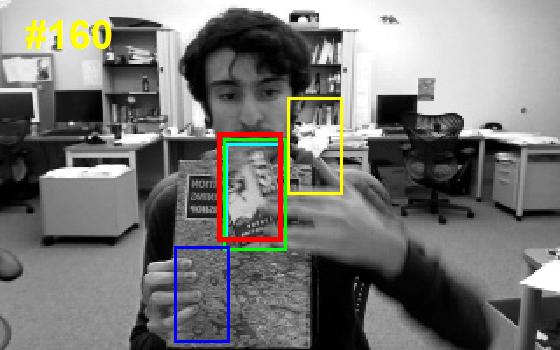}
\includegraphics[width=0.19\linewidth]{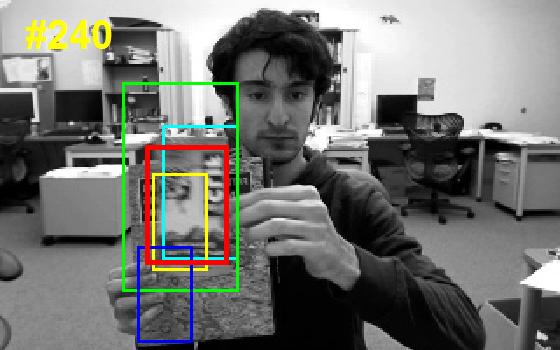}
\includegraphics[width=0.19\linewidth]{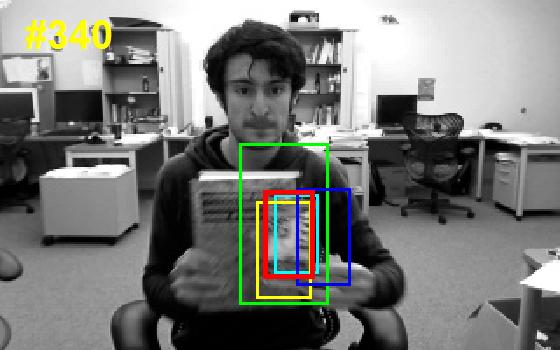}
\includegraphics[width=0.19\linewidth]{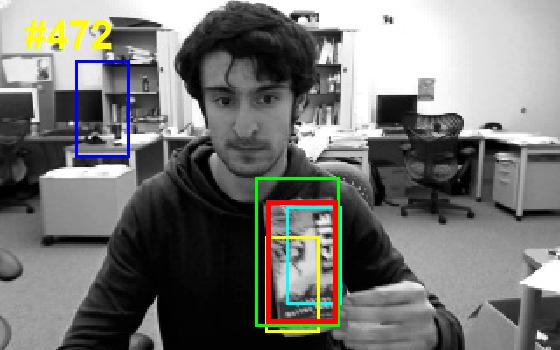}

\includegraphics[width=0.19\linewidth]{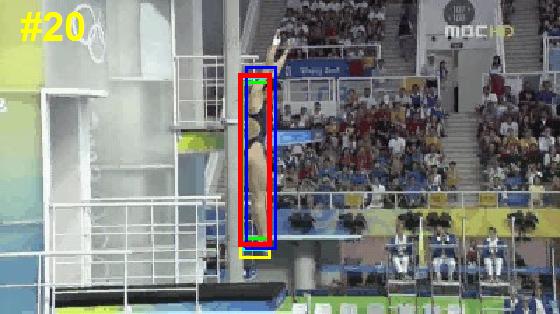}
\includegraphics[width=0.19\linewidth]{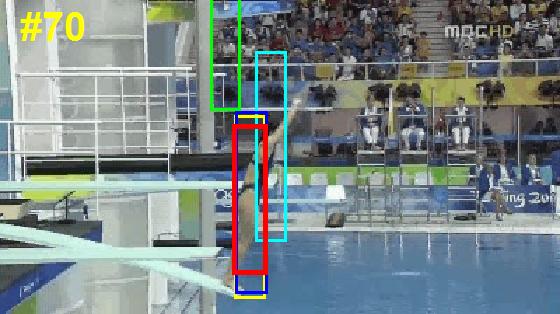}
\includegraphics[width=0.19\linewidth]{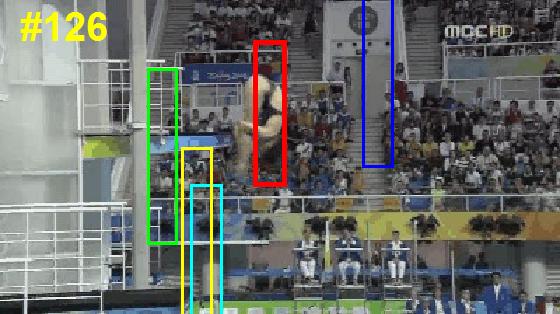}
\includegraphics[width=0.19\linewidth]{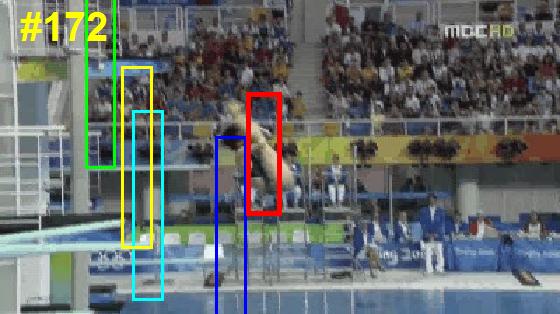}
\includegraphics[width=0.19\linewidth]{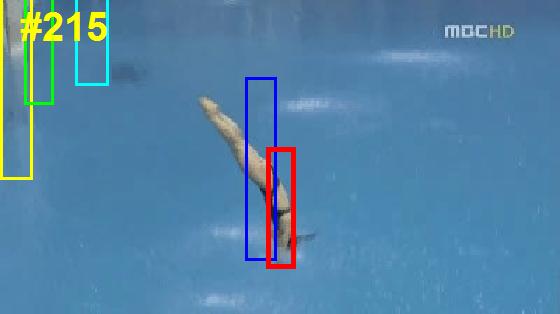}

\includegraphics[width=0.19\linewidth]{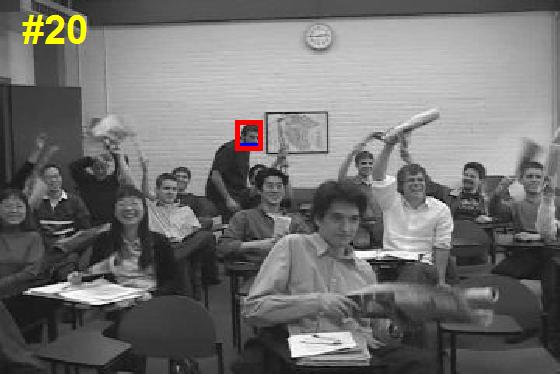}
\includegraphics[width=0.19\linewidth]{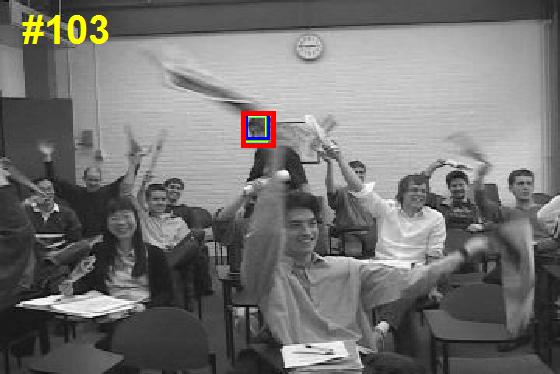}
\includegraphics[width=0.19\linewidth]{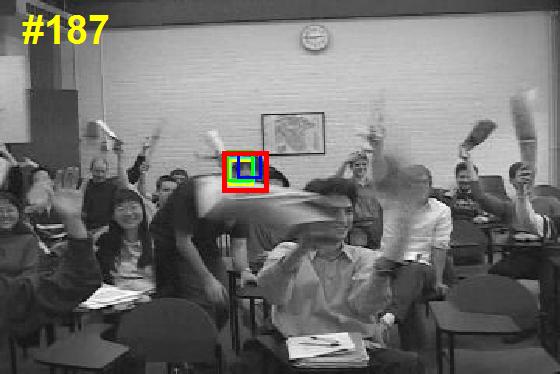}
\includegraphics[width=0.19\linewidth]{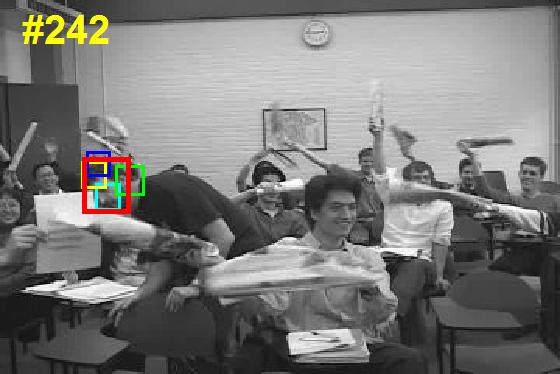}
\includegraphics[width=0.19\linewidth]{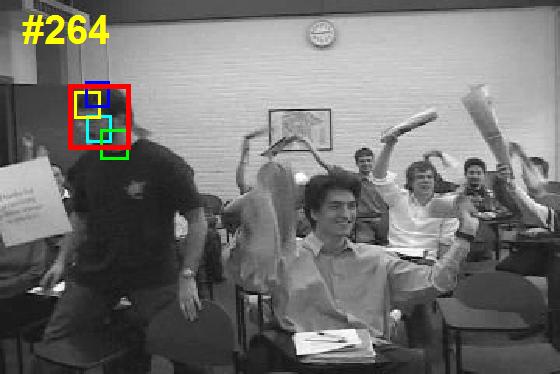}

\includegraphics[width=0.19\linewidth]{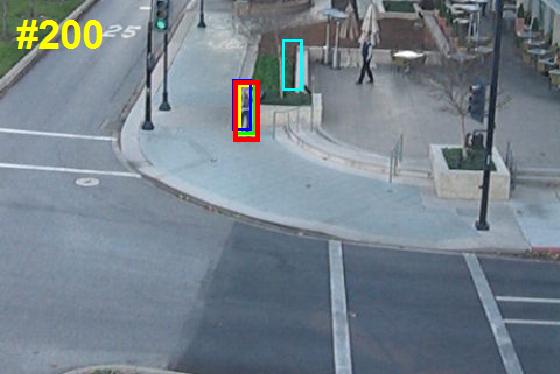}
\includegraphics[width=0.19\linewidth]{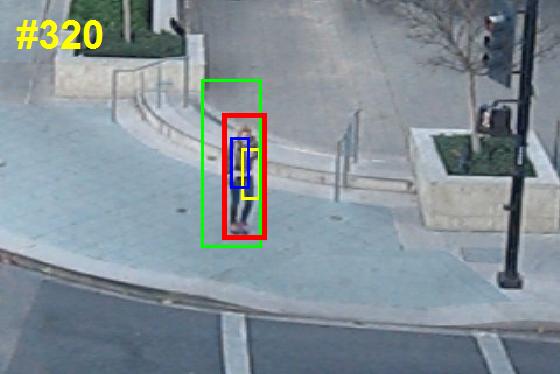}
\includegraphics[width=0.19\linewidth]{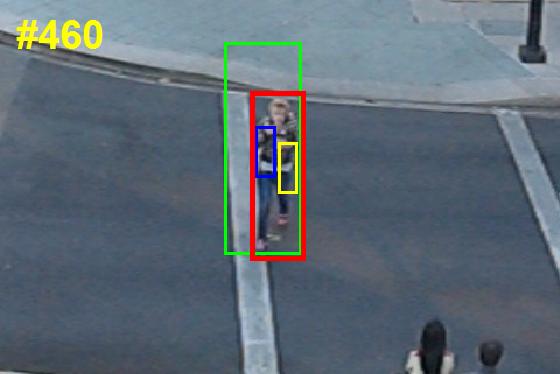}
\includegraphics[width=0.19\linewidth]{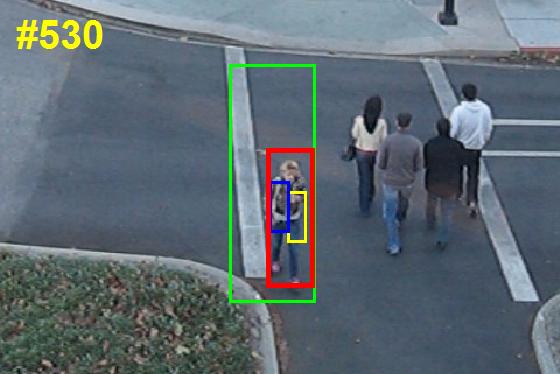}
\includegraphics[width=0.19\linewidth]{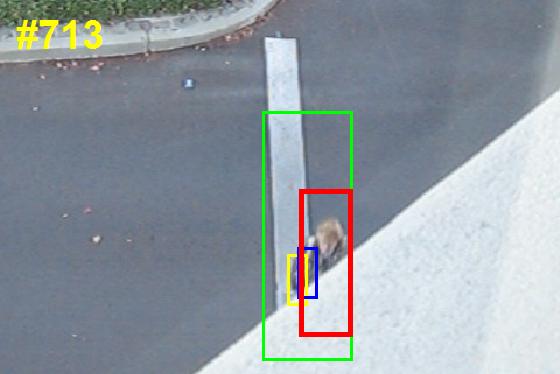}

\includegraphics[width=0.19\linewidth]{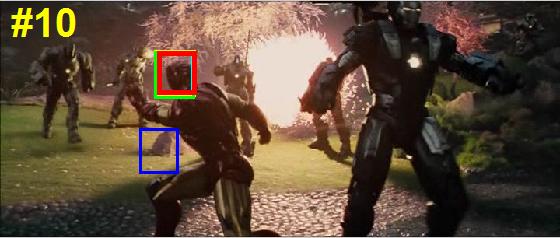}
\includegraphics[width=0.19\linewidth]{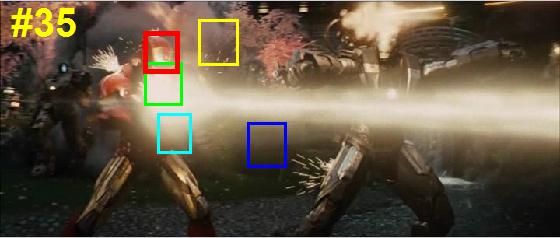}
\includegraphics[width=0.19\linewidth]{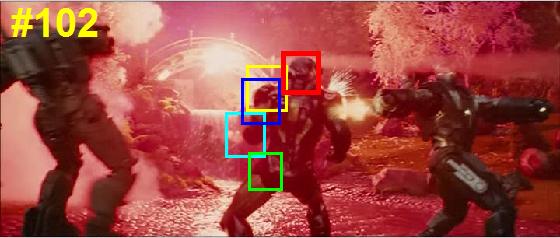}
\includegraphics[width=0.19\linewidth]{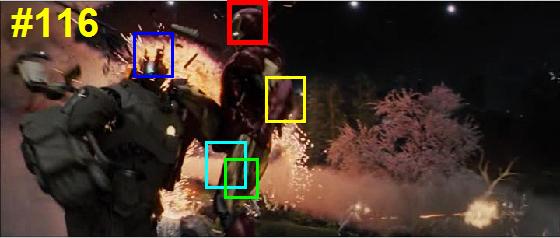}
\includegraphics[width=0.19\linewidth]{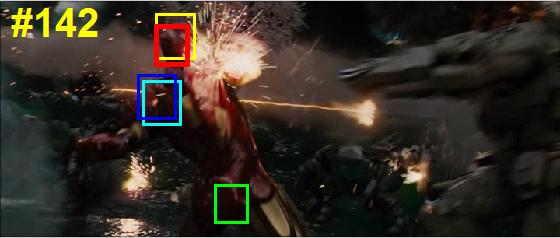}

\includegraphics[width=0.19\linewidth]{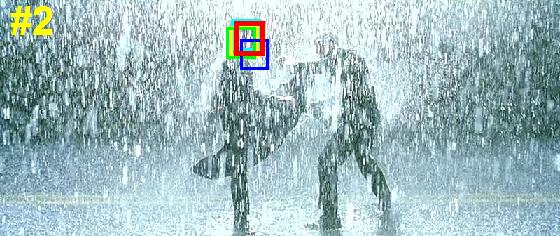}
\includegraphics[width=0.19\linewidth]{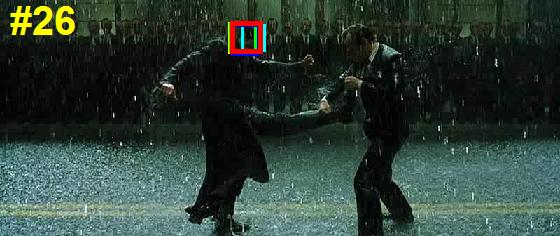}
\includegraphics[width=0.19\linewidth]{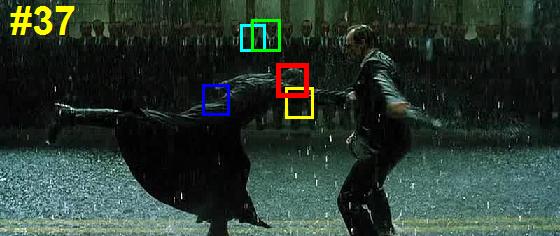}
\includegraphics[width=0.19\linewidth]{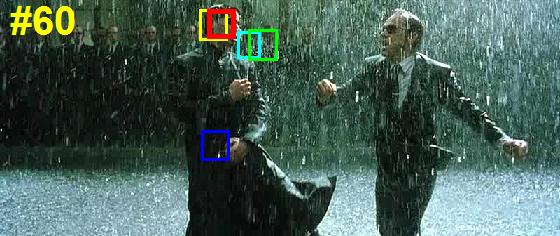}
\includegraphics[width=0.19\linewidth]{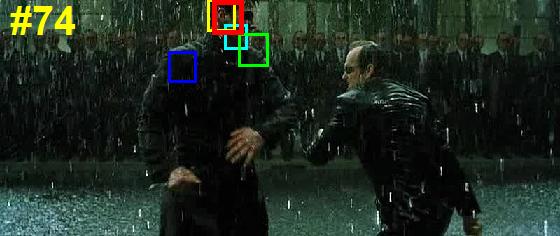}

\includegraphics[width=0.19\linewidth]{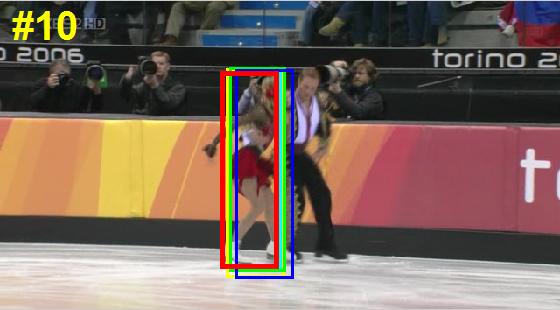}
\includegraphics[width=0.19\linewidth]{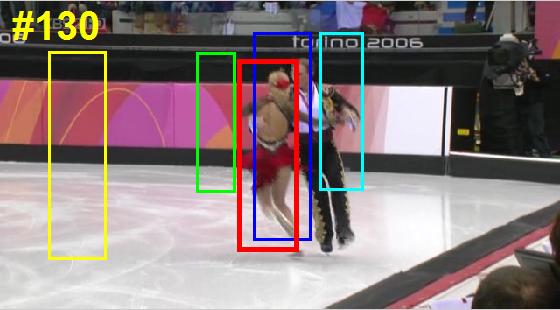}
\includegraphics[width=0.19\linewidth]{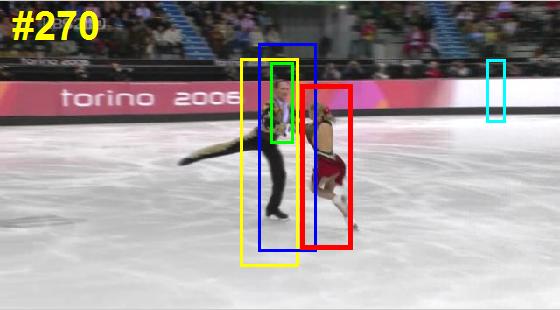}
\includegraphics[width=0.19\linewidth]{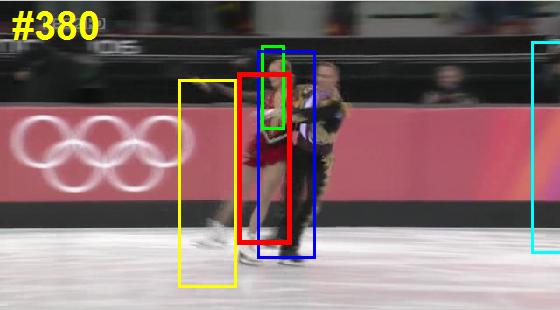}
\includegraphics[width=0.19\linewidth]{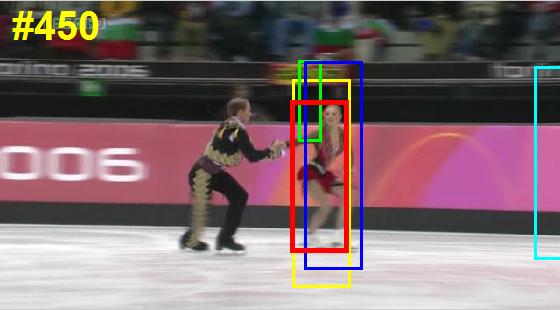}

\includegraphics[width=0.8\linewidth]{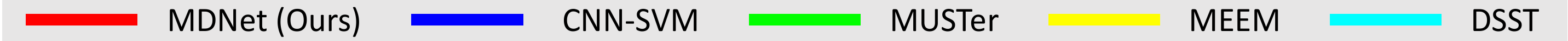}

\end{center}
\vspace{-5mm}
\caption{Qualitative results of the proposed method on some challenging sequences (\emph{Bolt2}, \emph{ClifBar}, \emph{Diving}, \emph{Freeman4}, \emph{Human5}, \emph{Ironman}, \emph{Matrix} and \emph{Skating2-1}).}
\label{fig:qual}
\end{figure*}

\begin{figure}[t]
\begin{center}
\includegraphics[width=0.32\linewidth]{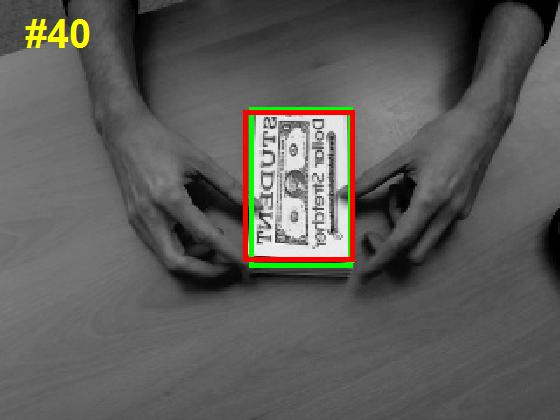}
\includegraphics[width=0.32\linewidth]{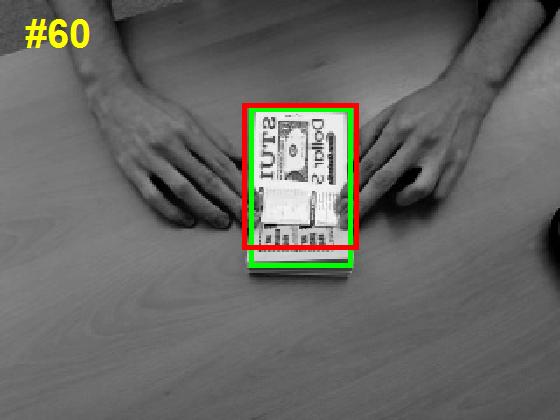}
\includegraphics[width=0.32\linewidth]{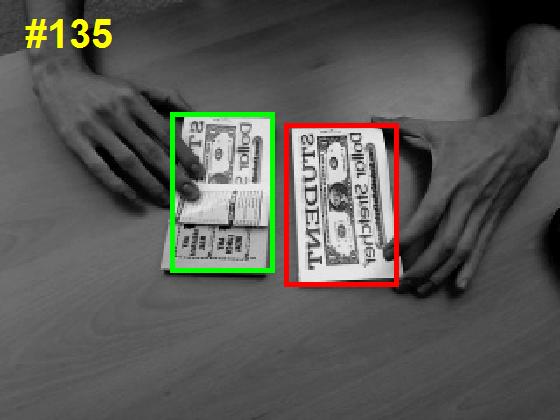}

\includegraphics[width=0.32\linewidth]{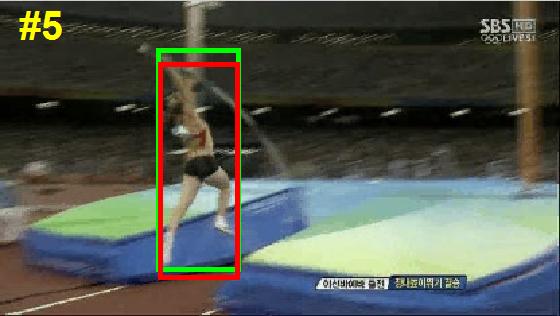}
\includegraphics[width=0.32\linewidth]{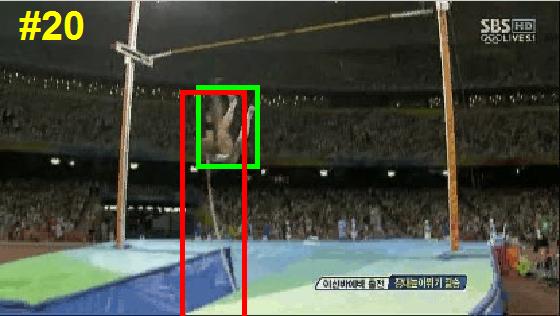}
\includegraphics[width=0.32\linewidth]{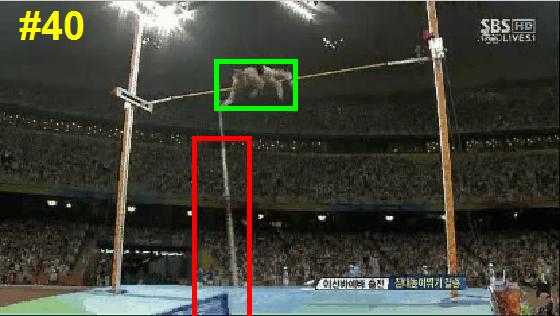}

\end{center}
\vspace{-5mm}
\caption{Failure cases of our method (\emph{Coupon} and \emph{Jump}). Green and red bounding boxes denote the ground-truths and our tracking results, respectively.}
\label{fig:failure}
\end{figure}

Figure~\ref{fig:qual} presents the superiority of our algorithm qualitatively compared to the state-of-the-art trackers.
Figure~\ref{fig:failure} shows a few failure cases of our algorithm; slight target appearance change causes a drift problem in \emph{Coupon} sequence, and dramatic appearance change makes our tracker miss the target completely in \emph{Jump} sequence.

\subsection{Evaluation on VOT2014 Dataset}

\begin{table*}
\small
\begin{center}
\begin{subtable}{0.48\textwidth}
\centering
\begin{tabular}{c|c c|c c||c}
\multirow{2}{*}{{Tracker}} 
& \multicolumn{2}{ c| }{{Accuracy}} 
& \multicolumn{2}{ c|| }{{Robustness}} 
& {Combined} \\
\cline{2-5}
& {Score} & {Rank}  & {Score} & {Rank} 
& {Rank}  \\\hline
{MUSTer}	& 0.58 & 4.50 & 0.99 & 5.67  & 5.09 \\
{MEEM}	& 0.48 & 7.17 & 0.71 & 5.50  & 6.34 \\
{DSST} 		& 0.60 & 4.03 & 0.68 & 5.17  & 4.60 \\
{SAMF}		& 0.60 & 3.97 & 0.77 & 5.58  & 4.78 \\
{KCF}		& {\color{blue}0.61} & {\color{blue}3.82} & 0.79 & 5.67  & 4.75 \\
{DGT} 		& 0.53 & 4.49 & 0.55 & 3.58  & {\color{blue}4.04} \\
{PLT\_14}	& 0.53 & 5.58 & {\color{red}0.14} & {\color{blue}2.75}  & 4.17 \\ [1pt] \hline
{MDNet}	& {\color{red}0.63} & {\color{red}2.50} & {\color{blue}0.16} & {\color{red}2.08}  & {\color{red}2.29} \\\hline
\end{tabular}
\vspace{-1mm}
\subcaption{Baseline result}
\end{subtable}
\begin{subtable}{0.48\textwidth}
\centering
\begin{tabular}{c|c c|c c||c}
\multirow{2}{*}{{Tracker}} 
& \multicolumn{2}{ c| }{{Accuracy}} 
& \multicolumn{2}{ c|| }{{Robustness}} 
& {Combined} \\
\cline{2-5}
& {Score} & {Rank}  & {Score} & {Rank} 
& {Rank}  \\\hline
{MUSTer}	& 0.55 & 4.67 & 0.94 & 5.53  & 5.10 \\
{MEEM}	& 0.48 & 7.25 & 0.74 & 5.76  & 6.51 \\
{DSST} 		& {\color{blue}0.58} & 4.00 & 0.76 & 5.10  & 4.55 \\
{SAMF}		& 0.57 & 3.72 & 0.81 & 4.94  & 4.33 \\
{KCF}		& {\color{blue}0.58} & 3.92 & 0.87 & 4.99  & 4.46 \\
{DGT} 		& 0.54 & {\color{blue}3.58} & 0.67 & 4.17  & 3.88 \\
{PLT\_14}	& 0.51 & 5.43 & {\color{red}0.16} & {\color{red}2.08}  & {\color{blue}3.76} \\ [1pt] \hline
{MDNet}	& {\color{red}0.60} & {\color{red}3.31} & {\color{blue}0.30} & {\color{blue}3.58}  & {\color{red}3.45} \\\hline
\end{tabular}
\vspace{-1mm}
\subcaption{Region\_noise result}
\end{subtable}
\end{center}
\vspace{-5mm}
\caption{The average scores and ranks of accuracy and robustness on the two experiments in VOT2014~\cite{vot14}. The first and second best scores are highlighted in red and blue colors, respectively.}
\label{tab:vot}
\end{table*}

\begin{figure}[t]
\begin{center}
\begin{subfigure}{\linewidth}
\includegraphics[width=\linewidth]{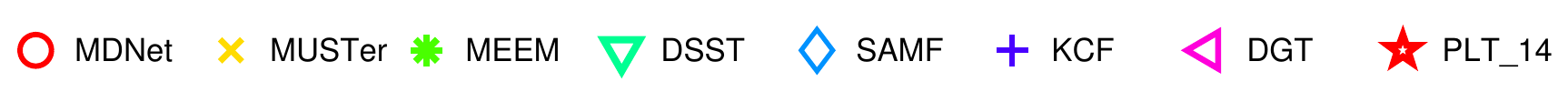}
\end{subfigure}
\begin{subfigure}{0.49\linewidth}
\includegraphics[width=\linewidth]{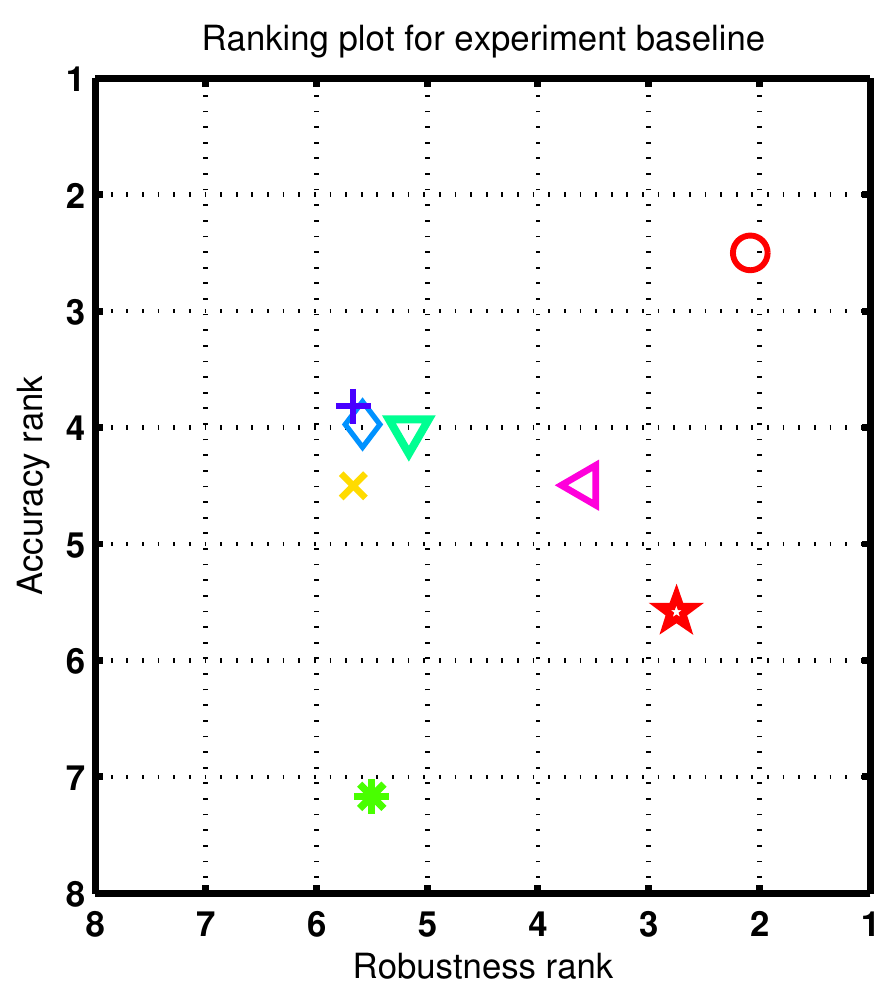}
\caption{Baseline}
\label{fig:otb50}
\end{subfigure}
\begin{subfigure}{0.49\linewidth}
\includegraphics[width=\linewidth]{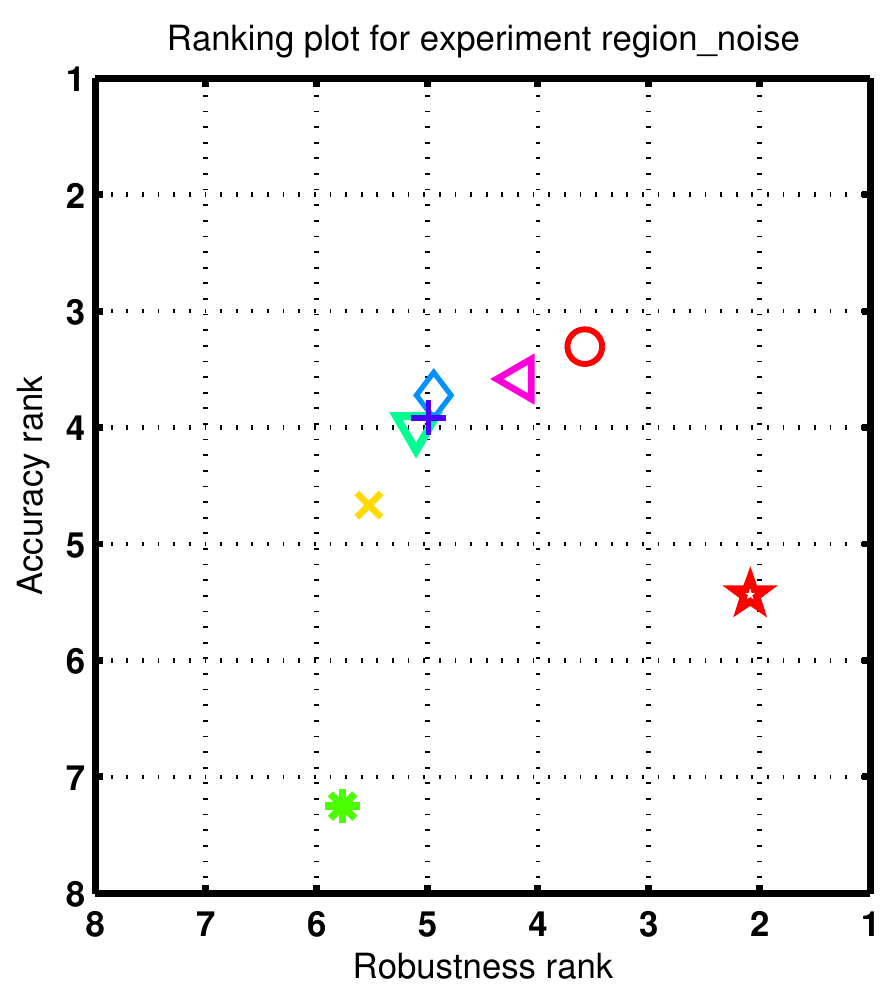}
\caption{Region\_noise}
\label{fig:otb100}
\end{subfigure}
\end{center}
\vspace{-5mm}
\caption{The robustness-accuracy ranking plots of tested algorithms in VOT2014 dataset. The better trackers are located at the upper-right corner.}
\label{fig:vot}
\end{figure}

\begin{figure}[t]
\vspace{-2mm}
\begin{center}
\includegraphics[width=\linewidth]{vot__tracker_legend-eps-converted-to.pdf}
\includegraphics[width=0.47\linewidth]{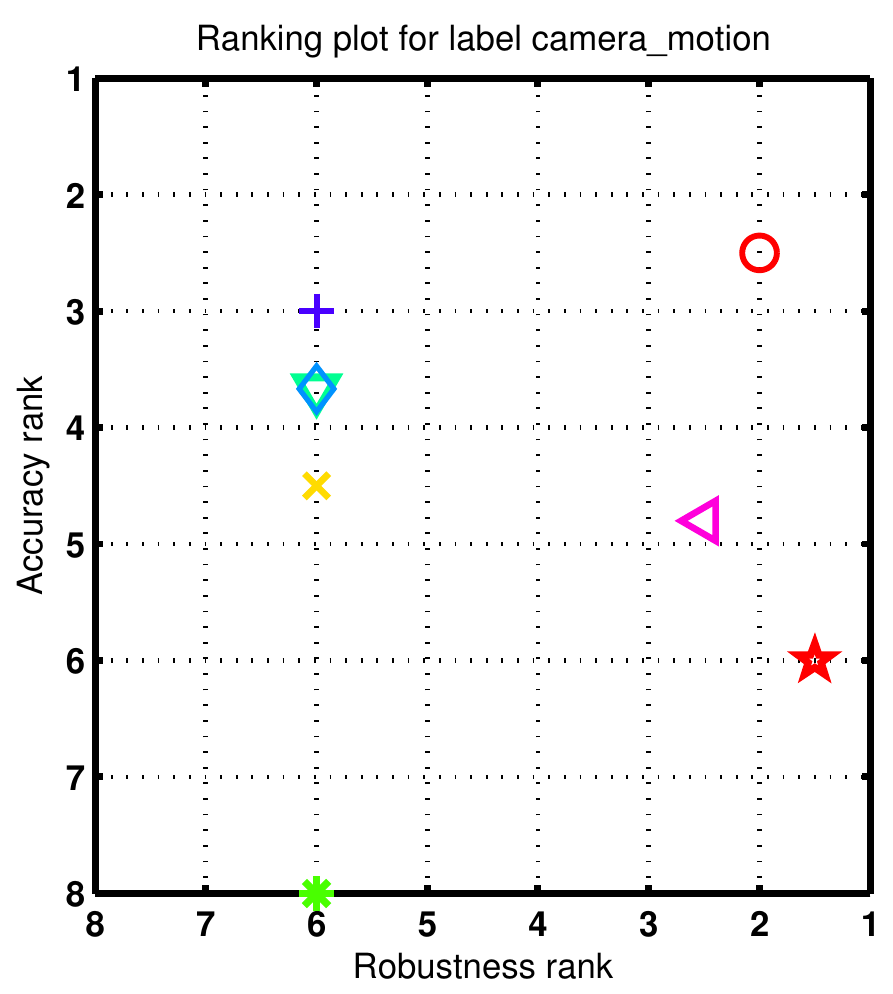}
\includegraphics[width=0.47\linewidth]{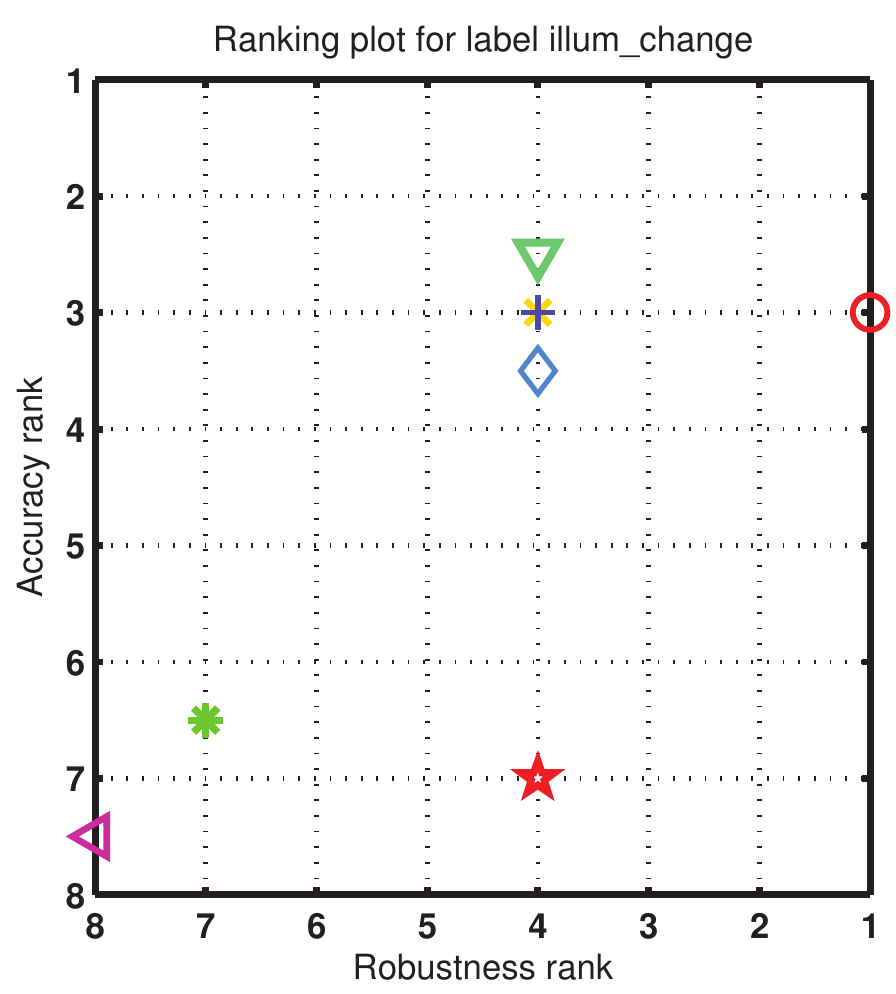}
\includegraphics[width=0.47\linewidth]{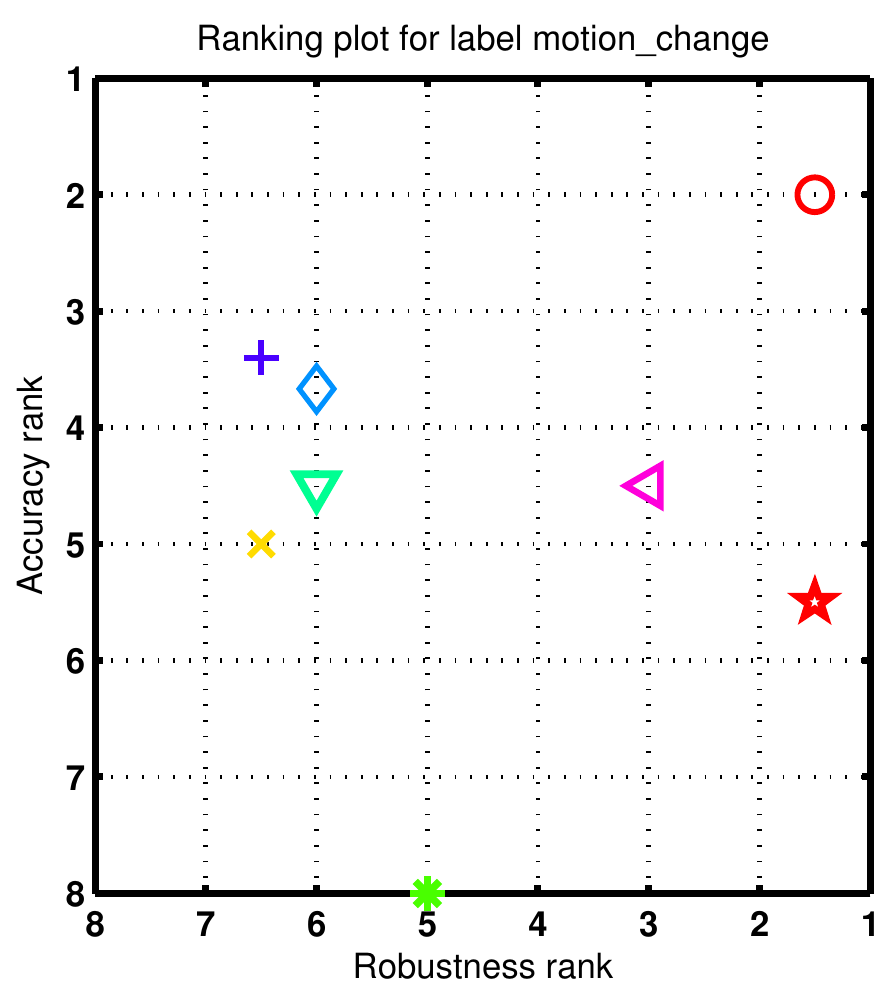}
\includegraphics[width=0.47\linewidth]{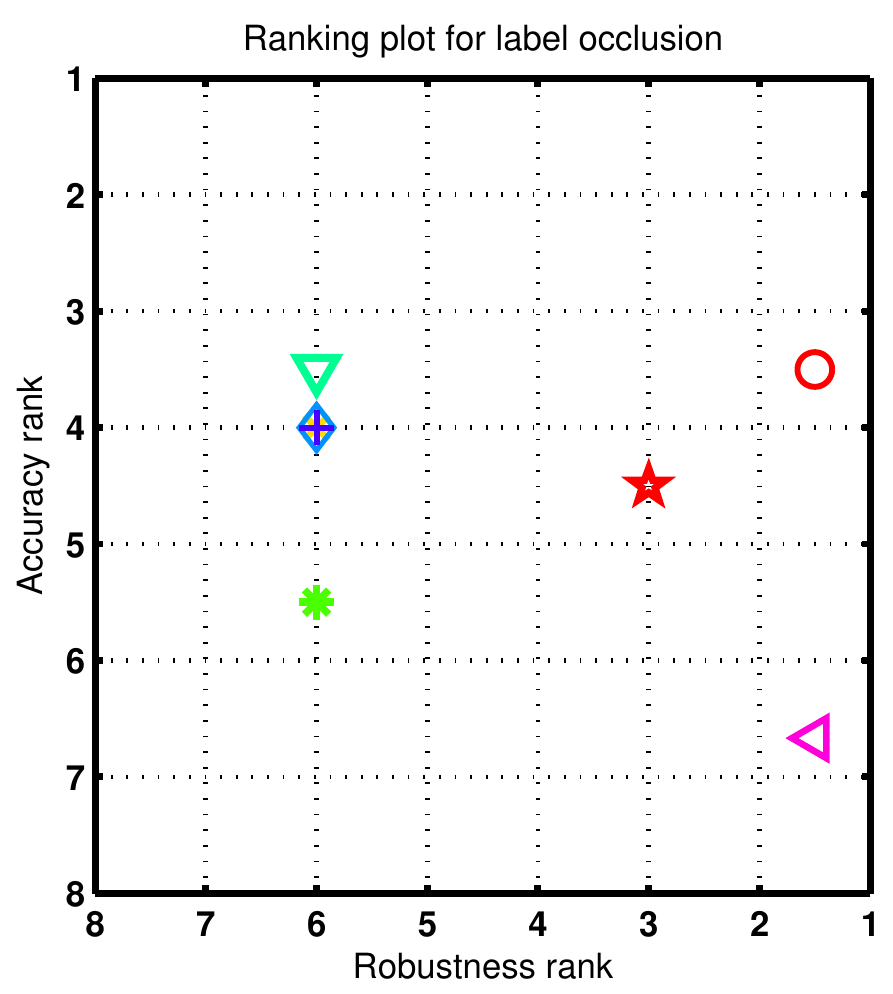}
\includegraphics[width=0.47\linewidth]{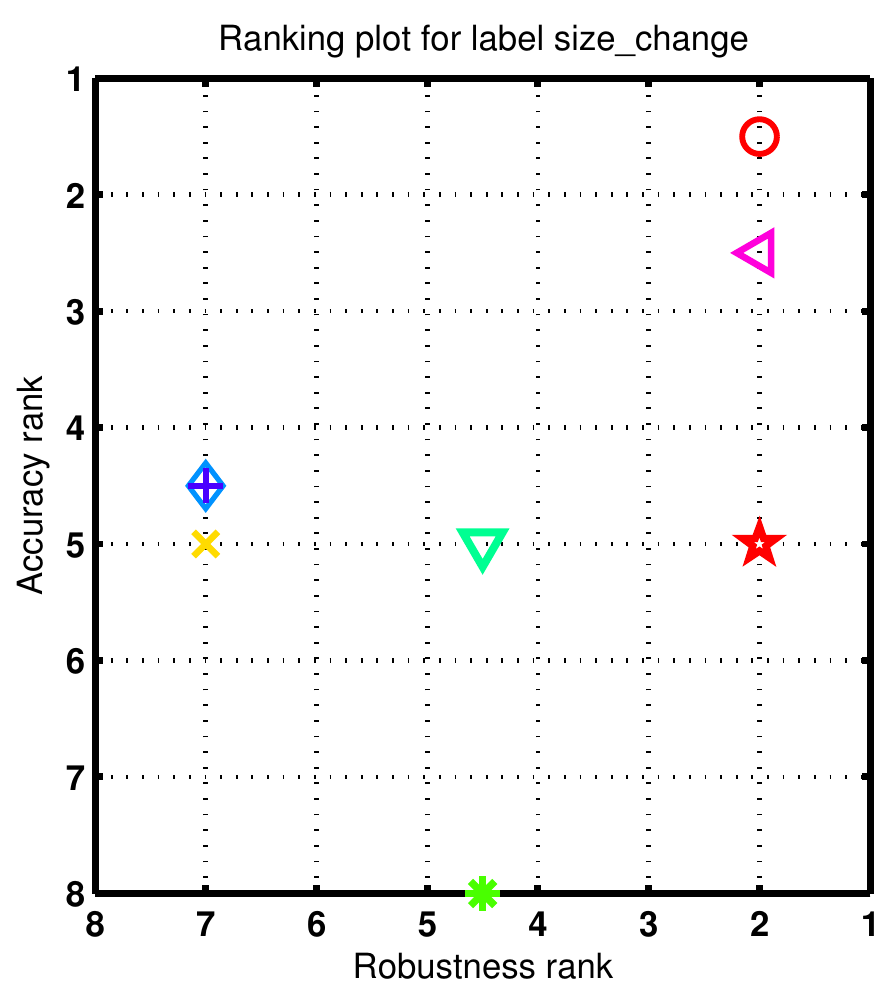}
\includegraphics[width=0.47\linewidth]{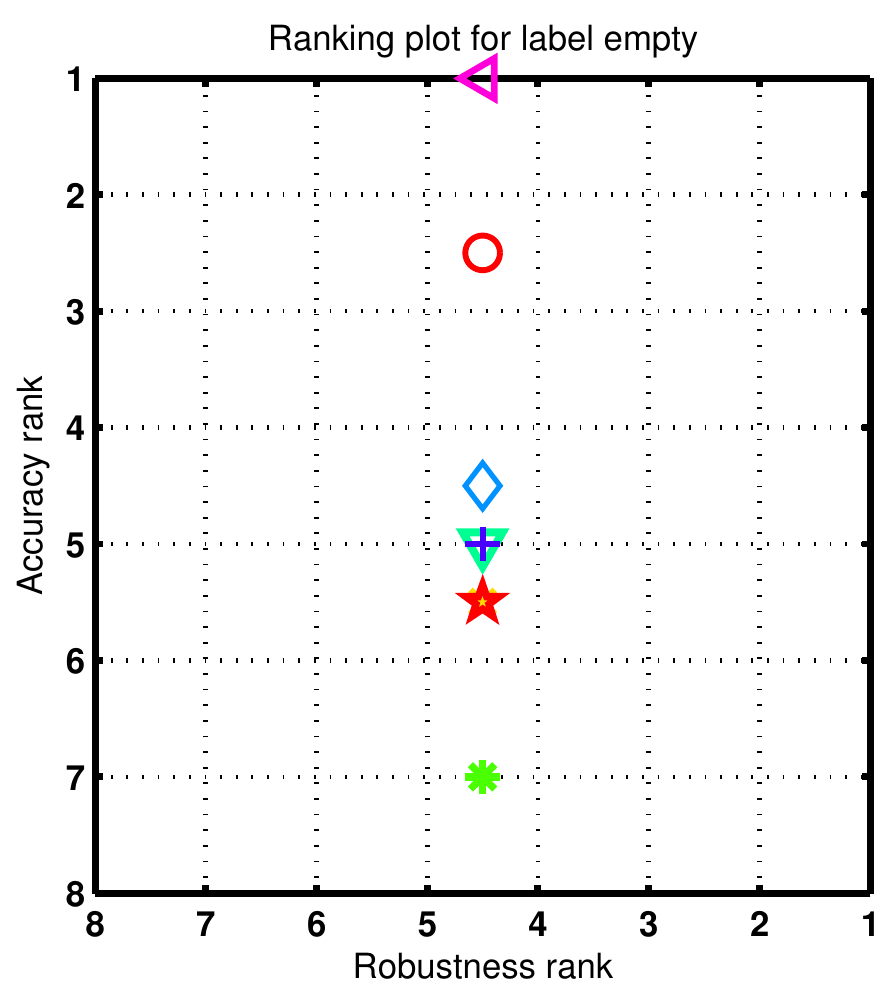}
\end{center}
\vspace{-5mm}
\caption{The robustness-accuracy ranking plots for five visual attributes: camera motion, illumination change, motion change, occlusion and  size change; and an empty attribute.}
\label{fig:vot_attributes}
\end{figure}

For completeness, we also present the evaluation results on VOT2014 dataset~\cite{vot14}, which contains 25 sequences with substantial variations. 
In the VOT challenge protocol, a tracker is re-initialized whenever tracking fails and the evaluation module reports both accuracy and robustness, which correspond to the bounding box overlap ratio and the number of failures, respectively.
There are two types of experiment settings; trackers are initialized with either ground-truth bounding boxes (baseline) or randomly perturbed ones (region\_noise).
The VOT evaluation also provides a ranking analysis based on both statistical and practical significance of the performance gap between trackers. 
Please refer to \cite{vot14} for more details.
We compare our algorithm with the top 5 trackers in VOT2014 challenge---DSST~\cite{danelljan2014accurate}, SAMF~\cite{li2014scale}, KCF~\cite{henriques2015high}, DGT~\cite{cai2013structured} and PLT\_14~\cite{vot14}---and additional two state-of-the-art trackers MUSTer~\cite{hong2015multi} and MEEM~\cite{zhang2014meem}.
Our network is pretrained using 89 sequences from OTB100, which do not include the common sequences with the VOT2014 dataset.
 
As illustrated in Table~\ref{tab:vot} and Figure~\ref{fig:vot}, MDNet is ranked top overall---the first place in accuracy and the first or second place in robustness; it demonstrates much better accuracy than all other methods, even with fewer re-initializations.
Furthermore, MDNet works well with imprecise re-initializations as shown in the region\_noise experiment results, which implies that it can be effectively combined with a re-detection module and achieve long-term tracking.
We also report the results with respect to several visual attributes from the baseline experiment in Figure~\ref{fig:vot_attributes}, which shows that our tracker is stable in various challenging situations.

\section{Conclusion}
\label{sec:conclusion}
We proposed a novel tracking algorithm based on a CNN trained in a multi-domain learning framework, which is referred to as MDNet.
Our tracking algorithm learns domain-independent representations from pretraining, and captures domain-specific information through online learning during tracking.
The proposed network has a simple architecture compared to the one designed for image classification tasks.
The entire network is pretrained offline, and the fully connected layers including a single domain-specific layer are fine-tuned online.
We achieved outstanding performance in two large public tracking benchmarks, OTB and VOT2014, compared to the state-of-the-art tracking algorithms.

{\small
\bibliographystyle{ieee}
\bibliography{mdnet}
}

\end{document}